\documentclass[10pt,twocolumn,letterpaper]{article}

\usepackage{cvpr}              % To produce the CAMERA-READY version
\definecolor{cvprblue}{rgb}{0.21,0.49,0.74}
\usepackage[pagebackref,breaklinks,colorlinks,allcolors=cvprblue]{hyperref}

\usepackage{makecell}
\usepackage{setspace}
\usepackage{verbatim}
\usepackage{amsmath}
\usepackage{diagbox}
\usepackage{multirow}
\usepackage[accsupp]{axessibility}
\def\paperID{} % *** Enter the Paper ID here 4376
\def\confName{CVPR}
\def\confYear{2026}

\title{Toward Early Quality Assessment of Text-to-Image Diffusion Models}

\author{
	Huanlei Guo$^{1}$\quad\quad Hongxin Wei$^{1}$\thanks{Corresponding authors: Hongxin Wei (weihx@sustech.edu.cn) and Bingyi Jing (bingyijing@cuhk.edu.cn).}\quad\quad Bingyi Jing$^{2,3}$\footnotemark[1]
	\\
	$^{1}$ Southern University of Science and Technology\\
	$^{2}$ The Chinese University of Hong Kong, Shenzhen~
	$^{3}$ Shenzhen Loop Area Institute % \\
}

\begin{document}
	\maketitle
	\begin{abstract}	
	Recent text-to-image (T2I) diffusion and flow-matching models can produce highly realistic images from natural language prompts. In practical scenarios, T2I systems are often run in a ``generate--then--select'' mode: many seeds are sampled and only a few images are kept for use. However, this pipeline is highly resource-intensive since each candidate requires tens to hundreds of denoising steps, and evaluation metrics such as CLIPScore and ImageReward are post-hoc. In this work, we address this inefficiency by introducing Probe-Select, a plug-in module that enables efficient evaluation of image quality within the generation process. We observe that certain intermediate denoiser activations, even at early timesteps, encode a stable coarse structure, object layout and spatial arrangement--that strongly correlates with final image fidelity. Probe-Select exploits this property by predicting final quality scores directly from early activations, allowing unpromising seeds to be terminated early. Across diffusion and flow-matching backbones, our experiments show that early evaluation at only 20\% of the trajectory accurately ranks candidate seeds and enables selective continuation. This strategy reduces sampling cost by over 60\% while improving the quality of the retained images, demonstrating that early structural signals can effectively guide selective generation without altering the underlying generative model. Code is available at \url{https://github.com/Guhuary/ProbeSelect}.
\end{abstract}

% 	In practical scenarios, text-to-image (T2I) systems are often run in a ``generate--then--select'' mode: many seeds are sampled and only a few images are kept for use. However, this pipeline is computationally expensive due to hundreds of iterative denoising steps per sample and selection is strictly post-hoc. With diffusion and flow matching models such as Stable Diffusion and Flux, selection is computationally expensive since common evaluators (e.g., CLIP scores and ImageReward) can score an image only after the full reverse process completes. This forces dozens of denoising steps per seed before any quality signal is available. We make a simple empirical observation: specific hidden states inside the denoiser change slowly over time and already contain a stable coarse layout that is predictive of the final image, even though the input latent is noisy. Built upon this observation, we introduce \emph{Probe-Select}, a plug-in module that predicts final-image quality from early intermediate states. The module reads a denoiser feature map with the normalized time $t$, and outputs a scalar score aligned to an external evaluator. During inference, we peek after a small fraction of steps and prune low-ranked seeds, reducing sampling cost while leaving the base sampler unchanged. Experiments on extensive T2I models show that early scoring can prune many seeds and keep the quality of the selected set comparable to late-only selection. 

	\section{Introduction}
Diffusion models, which leverage a progressive denoising process to synthesize data from noise, have established themselves as a cornerstone in generative AI \cite{ddpm, ncsn, song2020score}. Extending this framework, contemporary text-to-image systems, such as GLIDE \cite{glide}, DALL-E 2 \cite{DALLE2}, Stable Diffusion \cite{SD2, SD3} and Flux \cite{flux}, enable the creation of remarkably detailed and realistic images directly from textual descriptions, thereby facilitating widespread applications in creative industries and content production \cite{DALLE2, ding2023efficient, hartwig2025survey, bie2024renaissance}. In practical scenarios \cite{hartwig2025survey, bie2024renaissance}, users typically generate a large pool of candidate images per prompt and keep only a small subset under automatic evaluators such as ImageReward \cite{imagereward} and PickScore \cite{pickscore}. However, since diffusion sampling requires tens to hundreds of iterative steps per image and quality assessment is performed after the image is fully generated, substantial computation is devoted to candidates that may not meet the desired standards, representing a major computational bottleneck. This problem naturally leads to the task of early quality assessment (EQA), which aims to assess the potential quality of an emerging image trajectory after only a small fraction of the denoising steps, allowing for the timely interruption of less promising paths. 

Specialized quality metrics tailored for text-to-image generation, including CLIPScore \cite{clipscore}, ImageReward \cite{imagereward}, PickScore \cite{pickscore} and Human Preference Score \cite{hps}, have advanced considerably in capturing aspects like visual appeal, fidelity to the input prompt, and alignment with human preferences, thereby providing reliable tools for ranking completed outputs. Nevertheless, these evaluators operate exclusively on the final, fully generated images that emerge at the end of the denoising sequence, which means they depend on the completion of the entire iterative sampling procedure—a process that unfolds step by step and demands substantial processing for each candidate. To deal with this problem, recent work HEaD \cite{betti2024optimizing} has begun to explore intermediate signals (e.g., cross-attention maps) to predict object hallucinations and make binary continuation decisions in multi-object scenes. While promising, such task-specific detectors do not provide a general-purpose mechanism for forecasting overall image quality from partial generative states.

\begin{figure*}[!ht]
	\centering
	\includegraphics[width=0.88\textwidth]{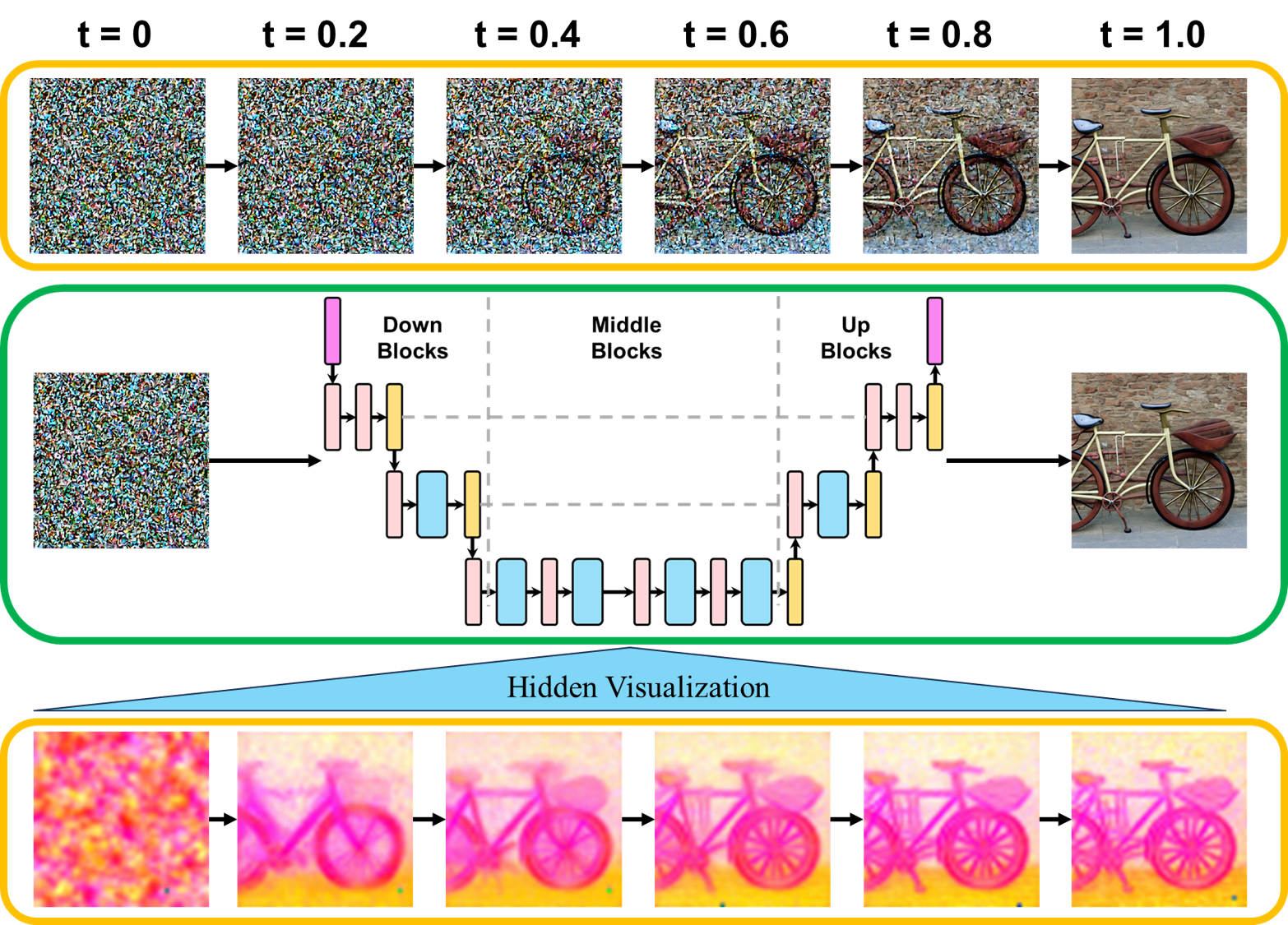}
	\caption{\emph{Top:} Snapshots of the reverse process in stable diffusion sampler from noisy latent to clean image. 
		\emph{Middle:} Denoiser architecture. 
		\emph{Bottom:} Visualization of hidden states. It shows that coarse layout and object contours emerge early and change slowly, which correlates with the final image.
	}
	\label{overview}
\end{figure*}

In this work, we bridge this gap by introducing \textbf{Probe-Select}, a plug-in framework for \textit{early quality assessment} in text-to-image diffusion models. 
Our key observation (Fig.~\ref{overview}) is that, even when the latent is still noisy, certain intermediate activations inside the denoiser already encode stable high-level structures—object layouts, spatial composition, and semantic groupings—that correlate strongly with the final image quality. 
Notably, these structural patterns appear early in the generation process and evolve slowly over time. 
This temporal stability provides a reliable basis for predicting final quality from partial generative states, thereby motivating our design of Probe-Select for early quality estimation. 

Concretely, we attach lightweight probes to selected denoiser features at an early checkpoint. By harnessing these structural cues and aligning probe outputs with external evaluators, Probe-Select can forecast eventual quality well before full denoising. During inference, Probe-Select runs the sampler to an early timestep and prunes low-scoring seeds, concentrating computation on promising candidates. This procedure concentrates computation on promising candidates with negligible overhead and without modifying the generator, sampler, or schedule and can be applied to various diffusion backbones.

Extensive experiments demonstrate the effectiveness of Probe-Select. First, on MS-COCO with five seeds per caption, probes attached at an early checkpoint (20\% of total sampling steps) already predict final quality with high fidelity—median Spearman correlations $\geq 0.7$ across eight evaluators, and $\approx 0.98-0.99$ for BLIP-ITM and ImageReward—remaining essentially stable across time. 
In addition, using these early scores to prune and continue only the top-1 of 5 candidates cuts expected denoising by $\approx 64 \%$ ( $\approx 0.36$ of full cost) while improving final quality across backbones. For Stable Diffusion 2 (SD2), ImageReward rises from 0.49 (baseline average over 5) to 1.59 with Probe-Select, and HPSv2.1 from 26.95 to 29.03. On Stable Diffusion 3.5 Medium (SD3-M) and Large (SD3-L), ImageReward reaches 1.83 with HPSv2.1 of 31.17 and 31.81 respectively. 

\begin{figure*}[!ht]
	\centering
	\includegraphics[width=0.85\textwidth]{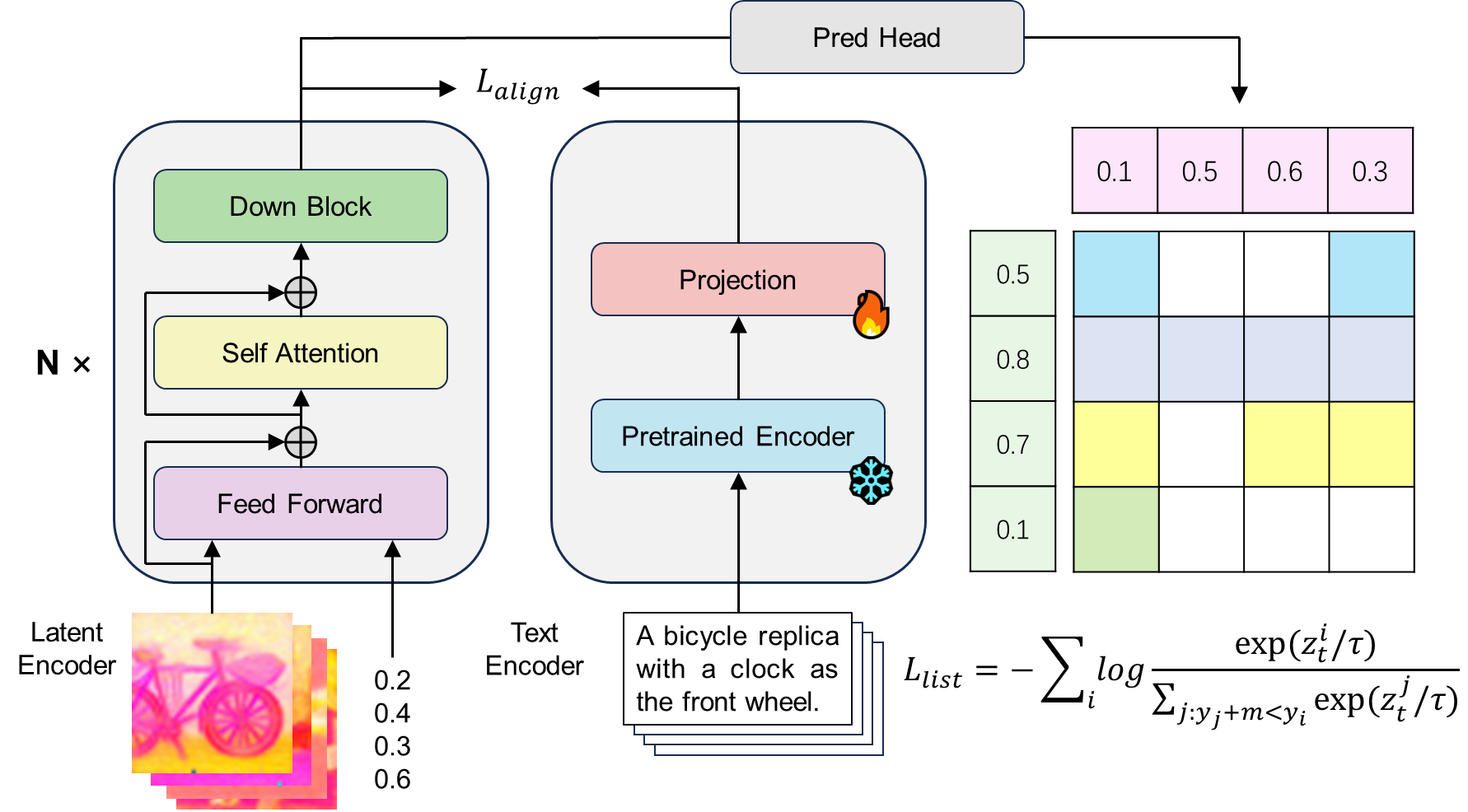}
	\caption{Overview of Probe-Select training. The model receive the intermediate denoiser activations and the timestep $t$ to produce the final quality. An additional text-aligned InfoNCE loss is employed for meaningful representation learning.
	}
	\label{pipeline}
\end{figure*}

Overall, Probe-Select plugs into various generative backbones without modifying the generator or its schedule, providing a general, compute-aware early assessment mechanism for scalable text-to-image generation. Our contributions can be summarized as follows:
\begin{itemize}
	\item \textbf{Early Assessment Paradigm:} We reframe text-to-image evaluation from a post-hoc task to a dynamic process that predicts quality from partial generative states.
	\item \textbf{Structural Signal Discovery:} We identify that stable structural cues in intermediate denoiser activations, which emerges as early as 20\% through the reverse process, can serve as reliable predictors of final image quality. 
	\item \textbf{Efficiency via Selective Generation:} We demonstrate that leveraging early predictions for trajectory pruning achieves substantial speedups with minimal quality trade-offs, generalizing across multiple generative backbones and evaluators.
\end{itemize}

	\section{Preliminaries: Generation-then-select}
We consider the image generation process of stable diffusion pipeline \cite{SD2, SD3}, which operates in the latent space of a pretrained autoencoder rather than the pixel domain. Let $D$ denotes the decoder of the autoencoder, $f_\theta$ be the denoiser conditioned on a text embedding $c(p)$ from prompt $p$. 
During inference, the generation process starts from a pure noise latent $z_{t=0}$ and progressively denoises it into the final latent $z_{t=1}$. 
Generation starts from a Gaussian latent $z_{t_0}$ at $t_0=0$ and proceeds on a time grid $0=t_0<t_1<\cdots<t_S=1$. The value $t$ denotes the normalized progress of the denoising process. 
At each step, the sampler queries $f_\theta$ and updates the latent according to a scheduler. We write the sampler in a scheduler-agnostic form:
\begin{equation}
	z_{t_{k+1}} = \Psi\!\left(z_{t_k}, t_k, t_{k+1}, f_\theta, c(p)\right),
\end{equation}
where $\Psi$ is the numerical stepper (e.g., DDIM, DPMSolver). After $S$ steps, the decoder reconstructs the final image $x_1$:
\begin{equation}
	x_1 = D(z_{t_S})=D(z_{t=1}).
\end{equation}

In practice, multiple candidates are sampled per prompt and ranked by external evaluators such as CLIPScore \cite{clipscore}, ImageReward \cite{imagereward}, or HPS \cite{hps}, which operate only on final images.
\begin{equation}
	y_m = R_m\left(x_1\right),
\end{equation}
where $R_m$ and $y_m$ denote different evaluator function and corresponding output value respectively. 

This \emph{post-hoc} evaluation requires completing full trajectory before scoring, wasting compute on low-quality candidates. Our goal is early quality assessment: predicting final quality from partial generative states so that only promising trajectories are continued, without changing the original diffusion model or schedule.

	\section{Proposed Method}
Our goal is to enable \textit{early quality assessment} in text-to-image generation—predicting the eventual perceptual quality of an image from partial generative information, long before the diffusion process finishes. 
Instead of waiting for complete denoising, we propose \textbf{Probe-Select}, a plug-in evaluator that attaches lightweight probes to intermediate denoiser activations and learns to forecast evaluator-aligned image quality. 
This early prediction supports multiple downstream applications selective generation or adaptive sampling, while serving as a general-purpose evaluation module. 
An overview of the training pipeline is shown in Fig.~\ref{pipeline}.

\subsection{Problem Formulation: Assessing Quality from Partial Generative States}
Diffusion generators gradually refine a noisy latent $z_0$ into a clean latent $z_1$ through a denoiser $f_\theta$ conditioned on the text embedding $c(p)$. 
Along this trajectory, the denoiser $f_\theta$ generates intermediate activations  $\mathbf{h}_{\mathbf{t}} \in \mathbb{R}^{C \times H \times W}$ at each timestep $t \in [0,1]$ from a selected block.
Existing evaluators assess the final image $x_1$ and we aim to infer similar quality signals from $h_t$ at an early timestep.

Formally, given an external evaluator $R_m(\cdot)$, we learn a predictor
\begin{equation}
	E_\phi(h_t, t) = p_\phi\big(g_\phi(h_t, t)\big) \rightarrow \hat y_{t, m},
\end{equation}
such that $\hat{y}_{t,m}\approx R_m(x_1)$. 
Here, $g_\phi$ encodes structural cues from $h_t$, and $p_\phi$ maps them to a scalar score. 
Once trained, $E_\phi$ provides an early approximation of external metrics, serving as a lightweight proxy for human- or reward-model preferences.

\subsection{Model Architecture: Early Structural Probes}
Although the latent $z_t$ is still noisy at early steps, certain denoiser's internal activations already reveal a coherent coarse structure—object layout, spatial composition, and emerging semantics—that changes slowly over time (Fig.~\ref{overview}). 
We leverage this property by attaching \emph{early structural probes} to $f_\theta$ to extract these signals. 

Probe-Select attaches \emph{early structural probes} to selected blocks of $f_\theta$ at an early checkpoint $t$ (e.g., $20\%$ steps):
\begin{itemize}
	\item \textbf{Feature taps.} We extract $h_{t}$ from chosen blocks in denoiser $f_\theta$. 
	\item \textbf{Probe encoder $g_\phi$.} A tiny vision encoder consumes $h_{t}$ and a timestep embedding, producing $u_{t}=g_\phi(h_{t}, t)\in\mathbb{R}^{d_h}$ with global pooling.
	\item \textbf{Projection head $p_\phi$.} A small MLP output a scalar $\hat y_{t}=p_\phi(u_{t})$.
\end{itemize}
The probe adds negligible overhead and requires no change to $f_\theta$ or the sampler, making the design backbone- and schedule-agnostic. 

\subsection{Training Objectives: Aligning with Evaluators and Prompts}
A naive regression from $u_t$ to evaluator scores is unstable and often ignores text semantics. To make early predictions both evaluator-consistent and prompt-sensitive, we combine two complementary objectives: (i) a listwise ranking loss that transfers the relative preference of evaluators, and (ii) a contrastive alignment loss that couples probe representations with prompt embeddings.  

\subsubsection*{Listwise Ranking for Preference Transfer}
Given a batch of $B$ samples with final evaluator scores $\left\{y_i\right\}$, we want early predictions $\left\{y_{t}^i\right\}$ to produce the same ranking. For notation simplicity, we omit the specific evaluator name $m$. 
A softmax-based listwise objective is applied as follows:
\begin{equation}
	\mathcal{L}_{\text{list}} = - \frac{1}{B} \sum_{i}^B \log \frac{\exp \big(\hat{y}_{t}^i / \tau_{\text{list}}\big)}{\sum_{j:,y_j + \alpha < y_i} \exp \big(\hat{y}_{t}^j / \tau_{\text{list}}\big)},
\end{equation}
where $\tau_{\text{list}} = \tau_{\text{list, max}} \times \max(1 - \frac{e}{E}, 0.1)$ is a temperature hyperparameter and $\alpha = \alpha_{\text{max}} \times (1 - \frac{e}{E})$ is a small margin. Here, $e$ and $E$ represent the current and max epoch during training, $\tau_{\text{list, max}}$ and $\alpha_{\text{max}}$ are the maximum value for temperature and margin. 

This loss emphasizes relative ordering rather than absolute values, encouraging the probe to focus on discriminative structural cues that separate good from bad seeds. 

\subsubsection*{Contrastive Text Alignment}
To maintain prompt awareness, we align the probe embedding $u_{t}$ with the prompt embedding $e_p= W_p E_{\text {text }}(p)$ extracted from a frozen text encoder (e.g., CLIP).
An InfoNCE loss encourages matched pairs ( $u_{t}, e_p$ ) to have high cosine similarity:
\begin{equation}
	\mathcal{L}_{\text{Align}} = -\frac{1}{B} \sum_{i=1}^B \log \frac{\exp\Big(\cos({u}_i, {e}_p^i) / \tau_{\text{Align}}\Big)}{\sum_{j=1}^B \exp\Big(\cos({u}_i, {e}_p^j) / \tau_{\text{Align}}\Big)},
\end{equation}
where $\tau_{\text{Align}} = \tau_{\text{Align, max}} \times \max(1 - \frac{e}{E}, 0.1)$ is the temperature. 

The total training loss is
\begin{equation}
	\mathcal{L}=\mathcal{L}_{\text {list }}+\lambda_{\text{Align}} \mathcal{L}_{\text{Align}}
\end{equation}
where $\lambda_{\text{Align}}$ balances ranking fidelity and semantic alignment.
Through this joint objective, the probe learns to produce text-aware quality estimates that are consistent with human or reward-model preferences-effectively serving as an early evaluator of generative quality.

\begin{figure*}[!ht]
	\centering
	% \footnotesize
	\newcommand{\tabincell}[2]{\begin{tabular}{@{}#1@{}}#2\end{tabular}}
	\begin{tabular}{m{0.6cm}<{\centering} m{1.6cm}<{\centering} m{1.6cm}<{\centering} m{1.6cm}<{\centering} m{1.6cm}<{\centering} m{1.6cm}<{\centering} m{1.6cm}<{\centering} m{1.6cm}<{\centering} m{1.6cm}<{\centering}}
		\tabincell{l}{Time}
		& \tabincell{c}{0.2}
		& \tabincell{c}{0.3}
		& \tabincell{c}{0.4}
		& \tabincell{c}{0.5}
		& \tabincell{c}{0.6}
		& \tabincell{c}{0.7}
		& \tabincell{c}{0.8}
		& \tabincell{c}{0.9}\\
		\rotatebox{90}{Down 1}
		& \includegraphics[width=18mm, height=18mm]{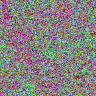}
		& \includegraphics[width=18mm, height=18mm]{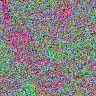}
		& \includegraphics[width=18mm, height=18mm]{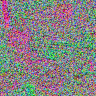}
		& \includegraphics[width=18mm, height=18mm]{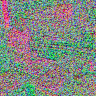}
		& \includegraphics[width=18mm, height=18mm]{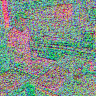}
		& \includegraphics[width=18mm, height=18mm]{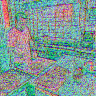}
		& \includegraphics[width=18mm, height=18mm]{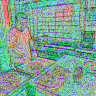}
		& \includegraphics[width=18mm, height=18mm]{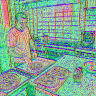} \\
		\rotatebox{90}{Down 2}
		& \includegraphics[width=18mm, height=18mm]{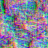}
		& \includegraphics[width=18mm, height=18mm]{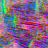}
		& \includegraphics[width=18mm, height=18mm]{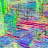}
		& \includegraphics[width=18mm, height=18mm]{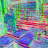}
		& \includegraphics[width=18mm, height=18mm]{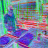}
		& \includegraphics[width=18mm, height=18mm]{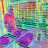}
		& \includegraphics[width=18mm, height=18mm]{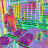}
		& \includegraphics[width=18mm, height=18mm]{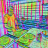} \\
		\rotatebox{90}{Up 3}
		& \includegraphics[width=18mm, height=18mm]{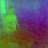}
		& \includegraphics[width=18mm, height=18mm]{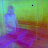}
		& \includegraphics[width=18mm, height=18mm]{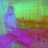}
		& \includegraphics[width=18mm, height=18mm]{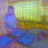}
		& \includegraphics[width=18mm, height=18mm]{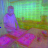}
		& \includegraphics[width=18mm, height=18mm]{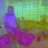}
		& \includegraphics[width=18mm, height=18mm]{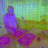}
		& \includegraphics[width=18mm, height=18mm]{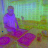} \\
		\rotatebox{90}{Up 4}
		& \includegraphics[width=18mm, height=18mm]{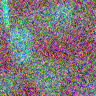}
		& \includegraphics[width=18mm, height=18mm]{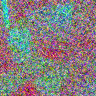}
		& \includegraphics[width=18mm, height=18mm]{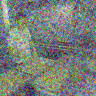}
		& \includegraphics[width=18mm, height=18mm]{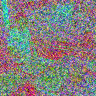}
		& \includegraphics[width=18mm, height=18mm]{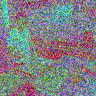}
		& \includegraphics[width=18mm, height=18mm]{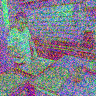}
		& \includegraphics[width=18mm, height=18mm]{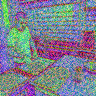}
		& \includegraphics[width=18mm, height=18mm]{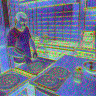} \\
	\end{tabular}
	\caption{PCA visualization for denoising network of Stable Diffusion 2 across time. }
	\label{fig: traj visual}
\end{figure*}

\subsection{Applications: Selective Sampling and Beyond}
Once trained, Probe-Select acts as a universal evaluator usable at any timestep. 
One practical application is selective sampling, where early predictions guide which trajectories to continue. 

Given $N$ random seeds for a prompt $p$, we run the model for a fraction $\eta$ of steps to obtain intermediate features $\left\{h_t^i\right\}$.
The probe scores them as $\{\hat{y}_t^i\}$, and only the top $K \ll N$ are continued to completion.
The expected computation cost is approximately
\begin{equation}
	\text { Cost Ratio } \approx \eta+(1-\eta) \frac{K}{N}
\end{equation}

This achieves large speedups without architectural changes.
The same early scores can further enable adaptive stopping (terminating once quality saturates) or quality-conditioned guidance adjustment, offering a general foundation for compute-aware text-to-image generation across diffusion-based frameworks.

	\section{Related Work}
\subsection{Efficiency in Diffusion Models}

One drawback of diffusion models is their slow generation speed. They often require many sampling steps to produce high-fidelity images \cite{song2020score, ddpm, ddim}, and each step is also expensive \cite{moon2024simple}. Most prior work improves efficiency by reducing the number of steps \cite{song2020score, luhman2021knowledge, lu2022dpm, salimans2022progressive, liu2022flow, flowmatching, liu2023instaflow, zhu2024slimflow, tang2023deediff, li2023autodiffusion}. For example, knowledge distillation transfers multi-step generation into fewer-step students \cite{luhman2021knowledge}, while flow matching-based methods encourage straighter sampling trajectories \cite{liu2022flow, flowmatching, liu2023instaflow}. DeeDiff \cite{tang2023deediff} uses uncertainty-aware early exiting, and AutoDiffusion \cite{li2023autodiffusion} searches for better time steps and architectures without retraining. Other work reduces per-step cost, such as latent diffusion in compressed spaces \cite{SD2, SD3} and adaptive block skipping \cite{moon2024simple}.

In this work, we take a complementary route: efficient evaluation. Instead of changing the sampler, we predict quality during sampling so that low-value seeds can be stopped early.

\subsection{Text-to-Image Evaluation}
% Traditional distributional metrics such as Inception Score \cite{IS} and Frechet Inception Distance \cite{FID} assess realism and diversity at the set level using features from a pretrained classifier. Alignment metrics based on vision–language encoders, for example CLIPScore \cite{clipscore} and BLIP-style models \cite{blip}, measure how well an image matches its prompt without a reference image. Preference-driven scorers, including ImageReward \cite{imagereward}, PickScore \cite{pickscore}, and Human Preference Score (HPSv2.0/HPSv2.1) \cite{hps}, ICTHP \cite{ICTHP}, learn from human comparisons to predict a scalar that correlates with user choices across styles and prompts. Aesthetic predictors trained on web-scale annotations provide an additional signal for visual appeal and are widely used for re-ranking \cite{as}. Despite their strengths, all these metrics operate on the final denoised image: they need the sampler to complete most or all steps and cannot consume noisy latents or intermediate activations. As a result, they offer no early signal for pruning seeds in diffusion or flow matching, which makes large-scale selection compute-intensive. In contrast, recent work on diffusion representations suggests that intermediate denoiser features already contain semantically meaningful structure well before the trajectory ends. This raises the possibility of using such early features not for external recognition tasks, but for online quality estimation during generation. 

Traditional distributional metrics like Inception Score \cite{IS} and Frechet Inception Distance \cite{FID} assess realism and diversity at the set level using features from a pretrained classifier. Vision–language encoder-based alignment metrics, such as CLIPScore \cite{clipscore} and BLIP-style models \cite{blip}, measure how well an image matches its prompt without a reference image. Preference-driven scorers, including ImageReward \cite{imagereward}, PickScore \cite{pickscore}, and Human Preference Score (HPSv2.0/HPSv2.1) \cite{hps}, learn from human comparisons to predict a scalar that correlates with user choices. Aesthetic predictors, trained on web-scale annotations, are used for re-ranking \cite{as}. However, all these metrics rely on final denoised images and cannot consume noisy latents or intermediate activations. This limits their ability to prune seeds early in diffusion or flow matching, making large-scale selection compute-intensive. Recent work on diffusion representations, however, suggests that intermediate denoiser features may contain meaningful structure before the trajectory ends, opening the possibility of using these features for online quality estimation during generation.

\subsection{Diffusion Features for Downstream Vision Tasks}
Recent studies show that pretrained diffusion models provide useful intermediate features for downstream tasks such as correspondence, retrieval, segmentation, and dense prediction \cite{dift, samuel2024waldo, zhang2024three, mukhopadhyay2024textfree}. Intermediate diffusion representations have also been reused to improve efficiency, controllability, and flexibility during generation \cite{deepcache, clockwork, becker2024controlling, plugandplay}. 

Unlike prior work, we show that certain diffusion U-Net activations contain meaningful semantic and structural information long before the final denoised image is formed. We leverage these early features for early quality assessment during sampling, enabling early stopping and candidate pruning.
	\begin{table*}[!ht]
	\centering
	\small
	\begin{tabular}{ccccccccc}
		\toprule
		Time & ClipScore & PickScore & AeS & BLIP-ITC & BLIP-ITM & ImageReward & HPSv2.0    & HPSV2.1    \\
		\midrule
		\multicolumn{9}{c}{SD2}\\
		\midrule
		0.1 & 0.54 & 0.62 & 0.51 & 0.52 & 0.71 & 0.70 & 0.53 & 0.49\\
		0.2 & 0.71 & 0.79 & 0.65 & 0.65 & 0.98 & 0.99 & 0.71 & 0.64\\
		0.3 & 0.71 & 0.79 & 0.65 & 0.65 & 0.98 & 0.99 & 0.71 & 0.64\\
		\midrule
		\multicolumn{9}{c}{SD3-M}\\
		\midrule
		0.1 & 0.59 & 0.67 & 0.53 & 0.60 & 0.70 & 0.71 & 0.61 & 0.62\\
		0.2 & 0.78 & 0.84 & 0.67 & 0.74 & 0.98 & 0.99 & 0.82 & 0.79\\
		0.3 & 0.78 & 0.84 & 0.67 & 0.74 & 0.98 & 0.99 & 0.82 & 0.79\\
		\midrule
		\multicolumn{9}{c}{SD3-L}\\
		\midrule
		0.1 & 0.64 & 0.68 & 0.57 & 0.62 & 0.76 & 0.78 & 0.62 & 0.61\\
		0.2 & 0.79 & 0.84 & 0.70 & 0.74 & 0.98 & 0.99 & 0.81 & 0.77\\
		0.3 & 0.79 & 0.84 & 0.70 & 0.74 & 0.98 & 0.99 & 0.82 & 0.78\\
		\midrule
		\multicolumn{9}{c}{FLUX.1-dev}\\
		\midrule
		0.1 & 0.60 & 0.67 & 0.54 & 0.59 & 0.69 & 0.71 & 0.59 & 0.62\\
		0.2 & 0.75 & 0.86 & 0.69 & 0.72 & 0.98 & 0.99 & 0.79 & 0.78\\
		0.3 & 0.75 & 0.86 & 0.69 & 0.72 & 0.98 & 0.99 & 0.79 & 0.78\\
		\bottomrule
		\end{tabular}
	\caption{Spearman correlation between predicted score and each evaluator from latent feature in different time stamps. The results demonstrate remarkable stability, with correlation scores remaining constant to two significant figures across time.}
	\label{tab: spearm corre}
\end{table*}

\section{Experimental Results}
In this section, extensive experiments are conducted to evaluate Probe-Select quantitatively and qualitatively. We first verify that intermediate denoiser activations contain stable structural cues by visual analysis and by measuring how early probe predictions correlate with final metrics across multiple backbones. Then, additional studies are conducted to verify the effectiveness Probe-Select by performing selective generation with early quality assessment. 

\subsection{Experimental Setup}
\noindent\textbf{Backbones and datasets.}
We evaluate several widely used diffusion-based generators: Stable Diffusion 2 (SD2) \cite{SD2}, Stable Diffusion 3.5 Medium (SD3-M), Stable Diffusion 3.5 Large (SD3-L) \cite{SD3} and FLUX.1-dev \cite{flux}. 
Training captions are drawn from the MS-COCO dataset (100k unique captions). For each caption, five distinct random seeds are sampled, producing 500k images per backbone.

\begin{table*}[!ht]
	\centering
	\small
	\begin{tabular}{ccccccccc}
		\toprule
		Time & ClipScore & PickScore & AeS & BLIP-ITC & BLIP-ITM & ImageReward & HPSv2.0    & HPSV2.1  \\
		\midrule
		SD2 & 31.95 & 21.82 & 5.26 & 0.47 & 0.77 & 0.49 & 28.64 & 26.95\\
		SD2-BIM & {\bf 33.96} & \underline{22.31} & \underline{5.30} & {\bf 0.51} & {\bf 0.99} & \underline{1.09} & \underline{29.34} & \underline{28.17}\\
		SD2-IR & \underline{33.50} & {\bf 22.43} & {\bf 5.37} & \underline{0.50} & \underline{0.96} & {\bf 1.59} & {\bf 29.56} & {\bf 29.03} \\
		\midrule
		SD3-M & 32.43 & 22.65 & \underline{5.80} & 0.49 & 0.89 & 1.12 & 29.68 & 29.64\\
		SD3-M-BIM & {\bf 34.56} & \underline{23.10} & 5.76 & {\bf 0.52} & {\bf 0.99} & \underline{1.53} & \underline{30.32} & \underline{30.60} \\
		SD3-M-IR & \underline{34.15} & {\bf 23.14} & {\bf 5.84} & {\bf 0.52} & \underline{0.98} & {\bf 1.83} & {\bf 30.50} & {\bf 31.17} \\
		\midrule
		SD3-L & 32.40 & 22.75 & \underline{5.91} & 0.49 & 0.89 & 1.14 & 30.00 & 30.29\\
		SD3-L-BIM & {\bf 34.43} & {\bf 23.28} & \underline{5.91} & {\bf 0.52} & {\bf 0.99} & \underline{1.56} & \underline{30.76} & \underline{31.42}\\
		SD3-L-IR & \underline{34.12} & \underline{23.26} & {\bf 5.98} & \underline{0.51} & \underline{0.98} & {\bf 1.83} & {\bf 30.85} & {\bf 31.81} \\ 
		\midrule
		FLUX.1-dev & 30.92 & 22.88 & 6.01 & 0.46 & 0.80 & 0.92 & 29.22 & 29.14\\
		FLUX.1-dev-BIM & \underline{33.03} & \underline{23.49} & \underline{6.07} & {\bf 0.51} & {\bf 0.99} & \underline{1.47} & \underline{30.22} & \underline{30.79}\\
		FLUX.1-dev-IR & {\bf 33.04} & {\bf 23.60} & {\bf 6.14} & \underline{0.50} & \underline{0.97} & {\bf 1.79} & {\bf 30.48} & {\bf 31.47}\\
		\bottomrule
	\end{tabular}
	\caption{Evaluation of models across various benchmarks. The best result is highlighted in bold, and the second best result is underlined. }
	\label{tab: selective generation}
\end{table*}

\noindent\textbf{Trajectory caching.}
Each model is executed under its default sampler and guidance settings.   
To enable temporal analysis, we cache denoiser activations at fractions of the generation schedule  
$t \in [0.2, 0.6]$ and store the final output at $t=1$.  
Each record therefore links a prompt–seed pair to its intermediate feature snapshots and final image. 
For each completed image we compute eight offline automatic metrics: CLIPScore \cite{clipscore}, PickScore \cite{pickscore}, AestheticScore (AeS) \cite{as}, BLIP-ITC \cite{blip}, BLIP-ITM \cite{blip}, ImageReward \cite{imagereward}, HPS v2.0 \cite{hps}, and HPS v2.1 \cite{hps}. These serve as the reference evaluators for training and analysis.

\noindent\textbf{Splits.}
Within each backbone’s 500k trajectories, 90\% are used for training and 10\% for validation. Testing uses MS-COCO validation captions, each with five random seeds. 

\noindent\textbf{Training Setup.}
Training runs up to $200$ epochs on $4$ NVIDIA A100-SXM4-40GB GPUs with batch size 480. We validate every epoch and monitor the validation spearman correlation score for early stopping with patience $20$. We use AdamW with learning rate $1e{-5}$ and weight decay $1e-2$. The schedule is cosine annealing with $\eta_{\min}=1e-6$. We maintain an exponential moving average of the probe parameters with decay $0.999$. We set $\lambda_{\text{Align}}=10$, $\tau_{\text{list, max}}=\tau_{\text{Align, max}}=1.0$, $\alpha_{\text{list, max}}=0.4\sigma$, where $\sigma$ is the standard deviation of target evaluator in training set. More detailed information is implemented in Appendix \ref{app:Probe-Select-impl}. 

\begin{figure}[ht]
	\centering
	\includegraphics[width=0.45\textwidth]{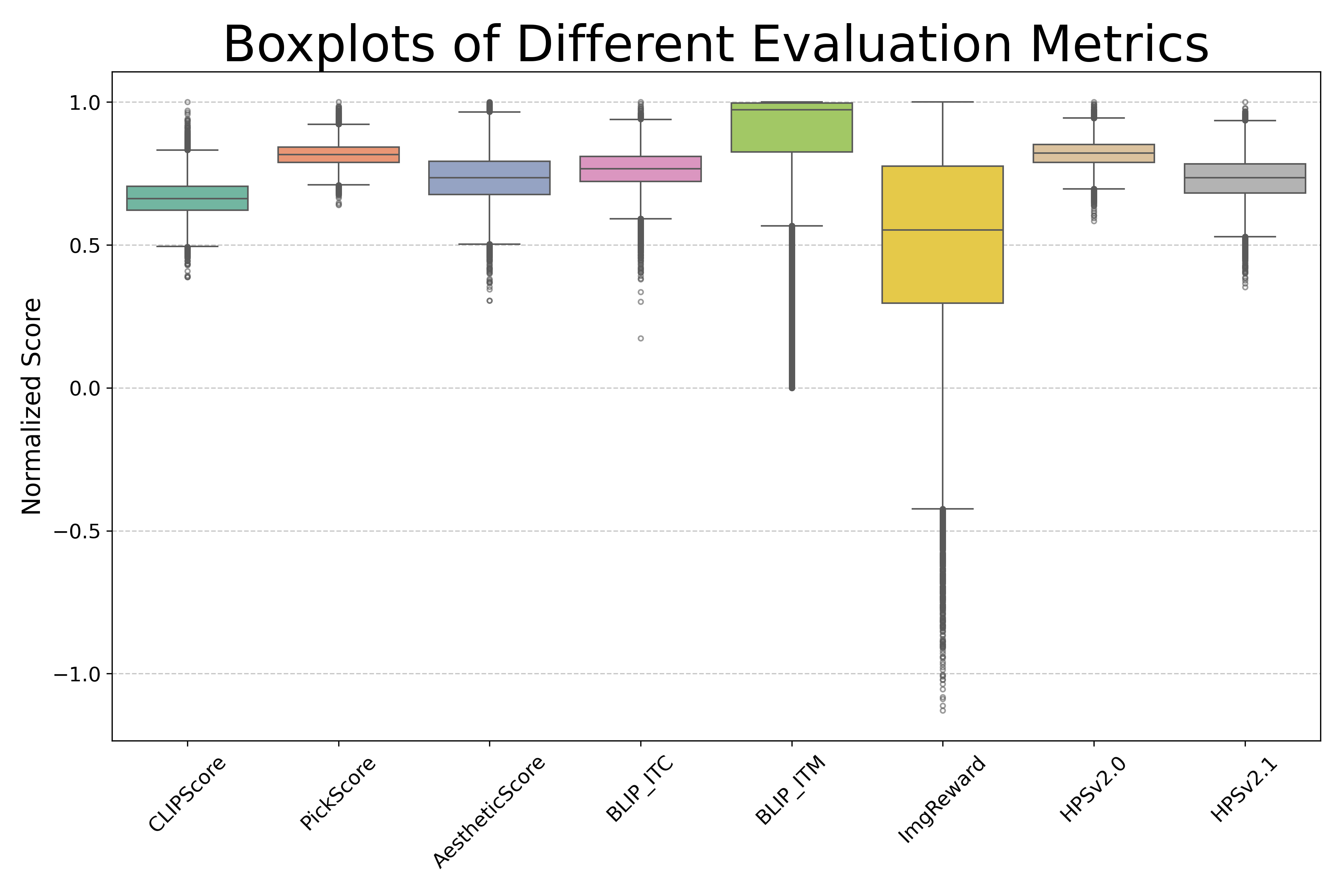}
	\caption{Boxplots of Evaluation Metrics of Samples Generated by SD3-L. All metric scores are normalized by dividing by their respective maximum values for visualization consistency. }
	\label{fig: box plots}
\end{figure}

\subsection{Early Structural Evidence in Denoiser Features}\label{sec: Early Structural Evidence in Denoiser Features}
Before quantitative evaluation, we visually inspect the evolution of several representative hidden activations of SD2 (output of 1st, 2nd down block and 3rd, 4th up block) across timesteps (Fig.~\ref{fig: traj visual}).
Additional visualizations of the full set of hidden states are provided in Appendix~\ref{appendix: additional results} for completeness.
Although the latents are still noisy in early steps, several mid-to-late layers of the denoiser already exhibit \textbf{clean and stable structural layouts}--object contours, spatial arrangement, and rough color distributions--that change only slowly thereafter. 

In SD2, the feature map from the 3rd upblock ({Up-3}) shows particularly strong stability: even when upstream inputs are highly corrupted, this layer preserves recognizable shapes and boundaries. 
This motivates our design choice to use this feature as the default ``probe tap'' for SD2. 
Repeating the analysis for SD3-M and SD3-L reveals analogous behavior in topologically similar layers, which is shown in Appendix \ref{appendix: additional results}. 
The consistency of early-emerging structural signals across backbones suggests that the model's internal representation already captures the essential layout long before completion—making early evaluation feasible.

\subsection{Quantitative Analysis: Predicting Final Quality from Partial States}\label{sec: Quantitative Analysis: Predicting Final Quality from Partial States}
We measure how well partial states forecast final quality by training probes at checkpoints $t\in\{0.1,0.2,0.3,0.4,0.5,0.6\}$ to predict each evaluator's final score and then computing Spearman correlation between the probe outputs and ground truth metrics. Table~\ref{tab: spearm corre} reports results for four backbones (SD2, SD3-M, SD3-L, and FLUX.1-dev). For completeness, results at $t=0.4, 0.5, 0.6$ are reported in the Appendix \ref{tab: full spearm corre}. We have following observations:

\noindent\textbf{High and stable correlations.}
Correlations are already strong at $t=0.2$ and change only slightly up to $t=0.6$. On SD2, most metrics are stable around $0.65$–$0.79$, while {BLIP-ITM} and {ImageReward} are near $1.0$ at all checkpoints. Larger models keep the same pattern with higher levels: on SD3-M, {HPSv2.1} rises from $0.79$ to $0.80$ as $t$ increases, and on SD3-L, {CLIPScore} and {AeS} (AestheticScore) hold around $0.80$ and $0.71$ from $t=0.4$ to $t=0.6$. The results on FLUX.1-dev shows similar stability, for example {PickScore} stays at $0.86$ and {HPSv2.1} stays at $0.78$ across all three checkpoints. 

These results indicate that by $t=0.2$ the features already contain enough information to rank seeds reliably, and later steps add only marginal gains for ranking. 

\noindent\textbf{Differences across evaluators.}
Correlation strength varies across metrics. BLIP-ITM and ImageReward reach values close to $1.0$, while CLIPScore, AestheticScore, and HPS v2.x typically plateau at lower levels. This difference is partly explained by what each evaluator emphasizes. Metrics such as ImageReward and BLIP-ITM rely more on holistic composition, object layout, and global semantic alignment, which are already largely formed in early denoiser features. By contrast, metrics such as CLIPScore and HPS are more sensitive to fine textures and high-frequency visual details that become reliable only later in the generation process. In addition, as shown in Fig.~\ref{fig: box plots}, evaluators with broader and less saturated score distributions (e.g., BLIP-ITM and ImageReward) induce more stable rank orderings, whereas narrower metrics create more ties and thus lower rank correlation. We leave the boxplots of SD2, SD3-M and FLUX.1-dev to Appendix~\ref{appendix: additional results}. 

Nevertheless, the overall consistency across metrics confirms that early activations embed a reliable signal predictive of multiple quality criteria. 
\subsection{Selective Generation with Early Assessment}
We use Probe-Select to rank partial trajectories at an early checkpoint and continue only the promising ones. For each caption, we draw five random seeds and run the generator only up to an early checkpoint at $t=0.2$ to extract intermediate denoiser features. A lightweight Probe-Select head then predicts the eventual {ImageReward} (IR) or {BLIP-ITM} (BIM) score for each partial trajectory. We continue only the top-1 seed ($K=1$, $N=5$) to full denoising and compute its final quality metrics. The {baseline} completes all five seeds with no early selection and reports the average metric, while the \emph{model–metric} variants (e.g., SD2-IR, SD3-M-BIM) select using the predicted metric and report the score of the chosen trajectory after full generation. 

\noindent\textbf{Main results across backbones.}
% Table \ref{tab: selective generation} reports results on four backbones: SD2, SD3-M, SD3-L, and FLUX.1-dev. On SD2, early selection guided by IR lifts {ImageReward} from $0.49$ (baseline) to $1.59$ (SD2-IR) and raises {HPSv2.1} from $26.95$ to $29.03$. The same trend holds on larger models: SD3-M-IR reaches $1.83$ IR and $31.17$ {HPSv2.1} (vs.\ $1.12$ and $29.64$ for the baseline), while SD3-L-IR attains the highest overall {HPSv2.1} at $31.81$ with $1.83$ IR. The FLUX.1-dev results show similar gains: IR rises from $0.92$ (baseline) to $1.79$ (FLUX.1-dev-IR), and {HPSv2.1} improves from $29.14$ to $31.47$. Perceptual metrics also improve under early selection, for example {ClipScore} on FLUX.1-dev increases from $30.92$ to $33.04$ (IR variant), and {PickScore} from $22.88$ to $23.60$. These patterns indicate that early predictions at $t=0.2$ align well with final image quality and human preference across different architectures. 
Table \ref{tab: selective generation} reports results on four backbones: SD2, SD3-M, SD3-L, and FLUX.1-dev. On SD2, early selection guided by IR improves {ImageReward} from $0.49$ to $1.59$ and {HPSv2.1} from $26.95$ to $29.03$. Similar gains appear on larger models: SD3-M-IR reaches $1.83$ IR and $31.17$ {HPSv2.1} (vs.\ $1.12$ and $29.64$ for the baseline), while SD3-L-IR achieves the highest overall {HPSv2.1} at $31.81$ with $1.83$ IR. FLUX.1-dev shows the same pattern, with IR increasing from $0.92$ to $1.79$ and {HPSv2.1} from $29.14$ to $31.47$. Perceptual metrics also improve: on FLUX.1-dev, {ClipScore} rises from $30.92$ to $33.04$, and {PickScore} from $22.88$ to $23.60$. Overall, these results suggest that early predictions at $t=0.2$ align well with final image quality and human preference across architectures.

We also observe modest gains on perceptual distributional quality. On MS-COCO, Probe-Select improves FID for both SD3-M ($25.26 \rightarrow 25.01$) and SD3-L ($23.72 \rightarrow 23.64$), indicating that early selection improves not only reward-based scores but also overall sample quality at the distribution level.

In terms of computational cost, this selective continuation reduces expected denoising cost by approximately 0.36, yielding about 64\% savings in sampling time while achieving better final quality. 

\begin{figure}[ht]
	\centering
	\includegraphics[width=0.45\textwidth]{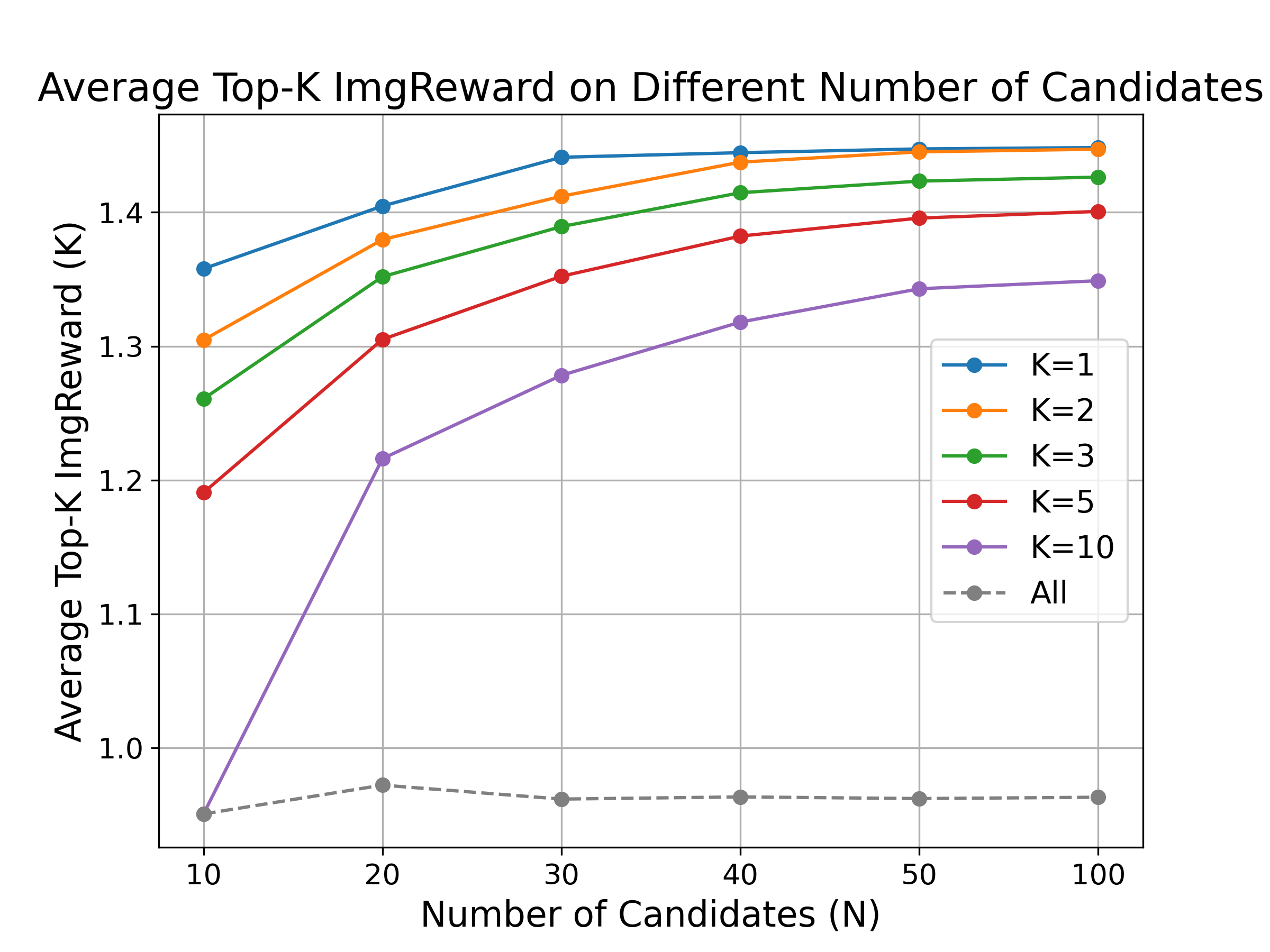}
	\caption{The relationship of number of candidates (N) and selected seeds (K). }
	\label{fig:nk_tradeoff}
\end{figure}

\noindent\textbf{Further study on $N$ and $K$.}
We further study the trade-off between the number of candidates $N$ and the number of continued seeds $K$ on FLUX.1-dev using $1000$ MS-COCO captions (Fig. \ref{fig:nk_tradeoff}). For each caption we draw $N\in{10,20,30,40,50,100}$ seeds, rank at $t_e=0.2$, and continue the top $K\in{1,2,3,5,10}$. We report the average Top-$K$ ImageReward; the dashed curve (“All”) averages over all $N$ without selection. Two patterns are consistent: (i) for fixed $K$, quality increases with $N$ and then saturates (Top-1 improves from about $1.36$ at $N=10$ to about $1.45$ at $N=100$, with most gains by $N\approx50$); (ii) for fixed $N$, the Top-$K$ average decreases as $K$ grows because lower-ranked samples are included. The no-selection baseline stays near $0.95$ regardless of $N$, showing that the improvement comes from early ranking rather than from drawing more candidates. When a single image is needed, $K=1$–$3$ with $N\approx50$ provides a good quality–cost balance.

These results demonstrate that early evaluation provides a reliable signal for ranking partial trajectories, enabling efficient ``generate-then-select'' workflows without sacrificing performance. Additional results for hyper-parameter tuning and visual analysis are presented in Appendix \ref{appendix: additional results}.
	\section{Conclusion}
We presented Probe-Select, a plug-in framework for efficient evaluation in text-to-image generation. By probing intermediate denoiser activations, our method predicts final image quality early in the generative process, bridging the gap between model-internal signals and post-hoc evaluators. Experiments on multiple diffusion backbones show that mid-to-late features encode stable structural cues strongly correlated with final quality.
Leveraging this property, Probe-Select enables selective generation—terminating low-quality trajectories early—achieving substantial computational savings with negligible quality loss.
Beyond acceleration, our findings highlight early evaluation as a general principle for adaptive and resource-efficient generative modeling.

Looking forward, we see several promising directions. Future work may extend early evaluation to dynamic timestep control, adaptive guidance, and cross-modal generation tasks.
Combining Probe-Select with reinforcement or active sampling frameworks could also enable closed-loop optimization, where evaluation directly steers generation in real time.
Ultimately, our findings highlight that understanding the evolution of internal representations during generation is not only diagnostically informative, but also operationally useful for building faster and more intelligent generative systems.
	\paragraph{Acknowledgments:} The authors acknowledge the financial support from the Natural Science Foundation of China (Grant No. 12371290) and the computational resources provided by the Center for Computational Science and Engineering of Southern University of Science and Technology. 
	{
		\small
		\bibliographystyle{ieeenat_fullname}
		\bibliography{ref}
	}
	
	\clearpage
\setcounter{page}{1}
\maketitlesupplementary
\begin{table*}[!ht]
	\centering
	\begin{tabular}{ccccccccc}
		\toprule
		Time & ClipScore & PickScore & AeS & BLIP-ITC & BLIP-ITM & ImageReward & HPSv2.0    & HPSV2.1    \\
		\midrule
		\multicolumn{9}{c}{SD2}\\
		\midrule
		0.4 & 0.71 & 0.79 & 0.65 & 0.65 & 0.98 & 0.99 & 0.71 & 0.64\\
		0.5 & 0.71 & 0.79 & 0.65 & 0.65 & 0.98 & 0.99 & 0.71 & 0.64\\
		0.6 & 0.71 & 0.79 & 0.65 & 0.65 & 0.98 & 0.99 & 0.71 & 0.64\\
		\midrule
		\multicolumn{9}{c}{SD3-M}\\
		\midrule
		0.4 & 0.78 & 0.84 & 0.67 & 0.74 & 0.98 & 0.99 & 0.82 & 0.80\\
		0.5 & 0.78 & 0.84 & 0.67 & 0.74 & 0.98 & 0.99 & 0.82 & 0.80\\
		0.6 & 0.78 & 0.84 & 0.67 & 0.74 & 0.98 & 0.99 & 0.83 & 0.80\\
		\midrule
		\multicolumn{9}{c}{SD3-L}\\
		\midrule
		0.4 & 0.80 & 0.85 & 0.71 & 0.74 & 0.98 & 0.99 & 0.82 & 0.78\\
		0.5 & 0.80 & 0.85 & 0.71 & 0.74 & 0.98 & 0.99 & 0.82 & 0.78\\
		0.6 & 0.80 & 0.85 & 0.71 & 0.74 & 0.98 & 0.99 & 0.82 & 0.78\\
		\midrule
		\multicolumn{9}{c}{FLUX.1-dev}\\
		\midrule
		0.4 & 0.75 & 0.86 & 0.69 & 0.72 & 0.98 & 0.99 & 0.79 & 0.78\\
		0.5 & 0.75 & 0.86 & 0.69 & 0.72 & 0.98 & 0.99 & 0.79 & 0.78\\
		0.6 & 0.75 & 0.86 & 0.69 & 0.72 & 0.98 & 0.99 & 0.79 & 0.78\\
		\bottomrule
	\end{tabular}
	\caption{Spearman correlation between predicted score and each evaluator from latent feature in different time stamps. }
	\label{tab: full spearm corre}
\end{table*}

\begin{figure*}[!ht]
	\centering
	\begin{subfigure}{0.3\textwidth}
		\includegraphics[width=\textwidth]{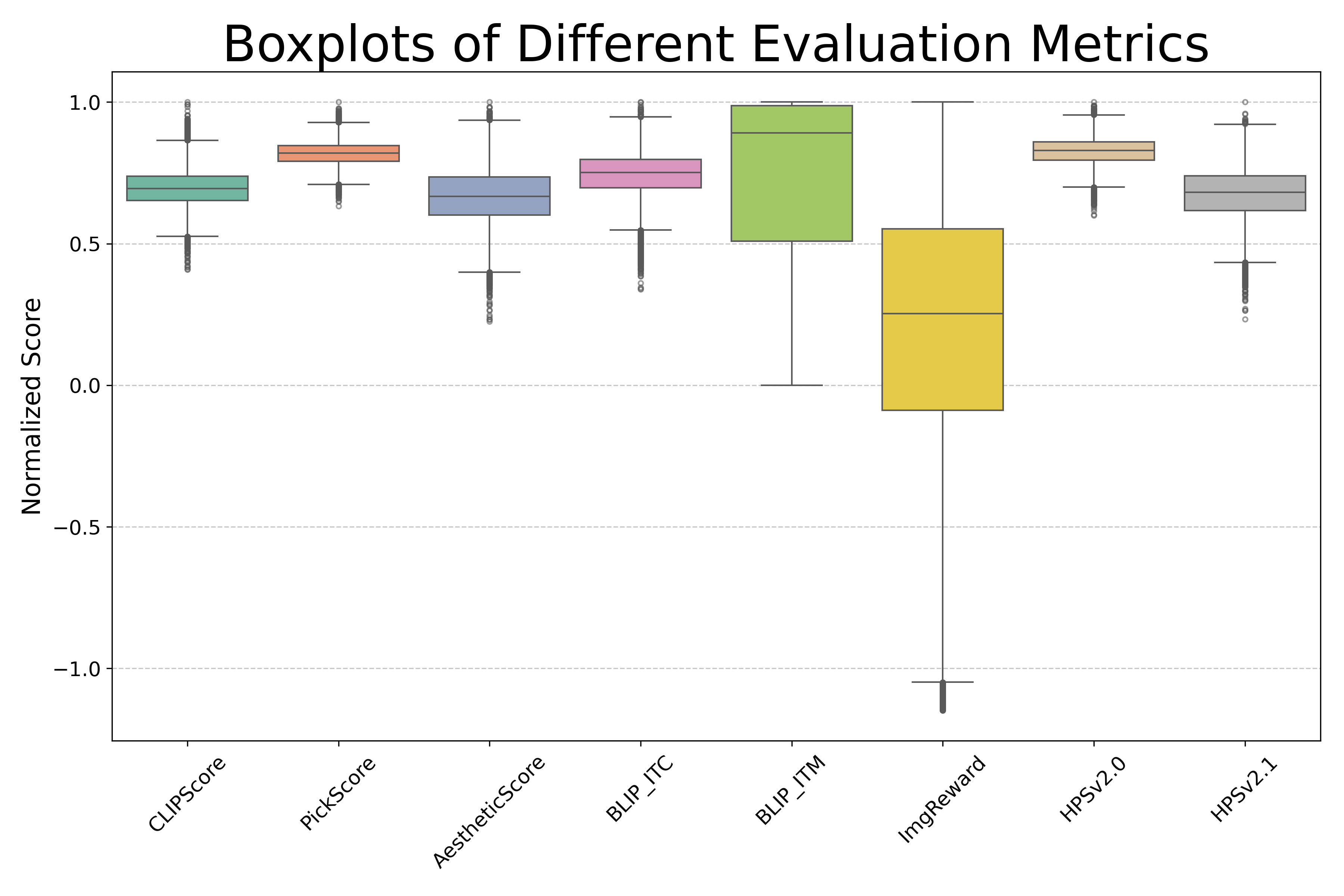}
		\caption{SD2}
	\end{subfigure}
	\begin{subfigure}{0.3\textwidth}
		\includegraphics[width=\textwidth]{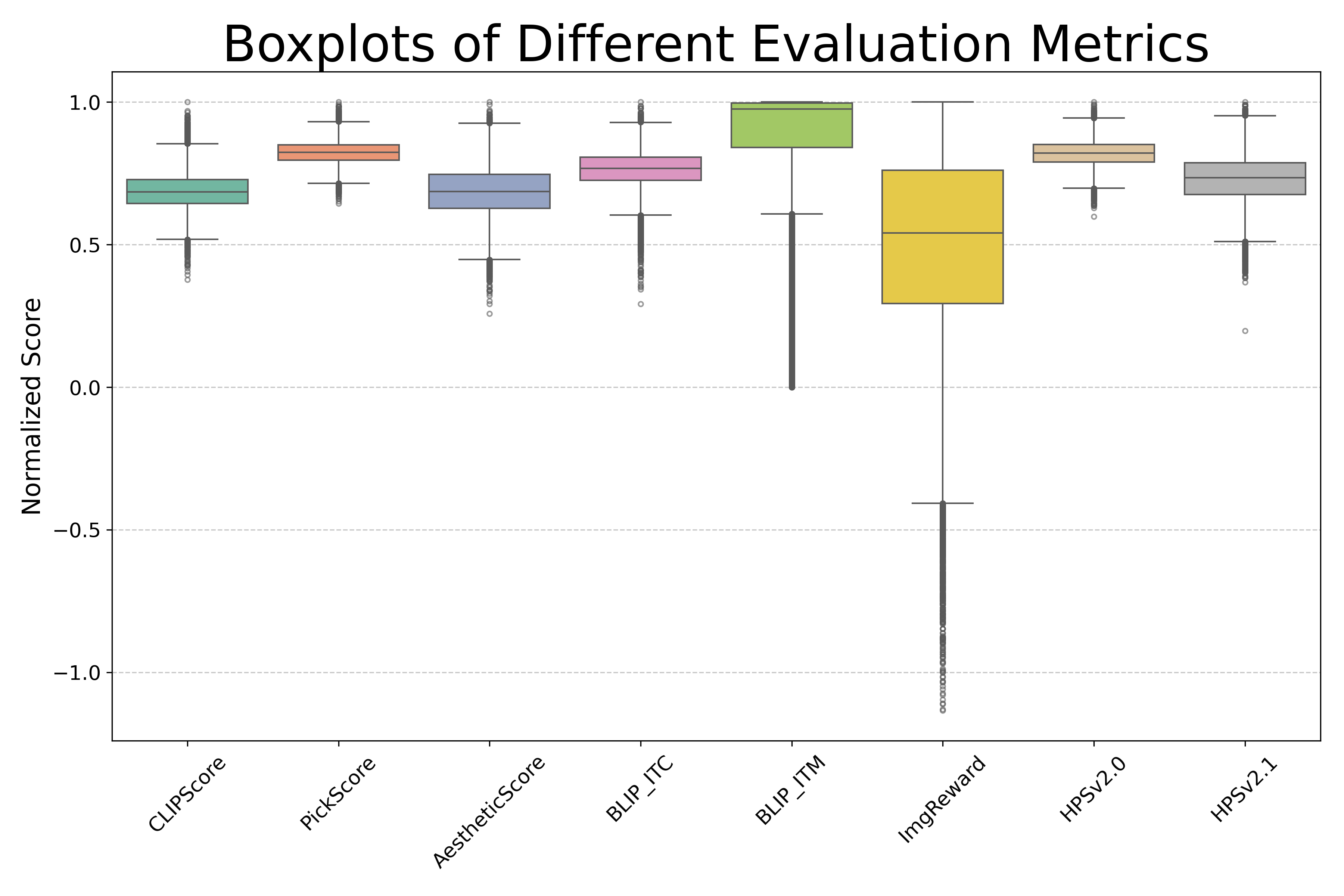}
		\caption{SD3M}
	\end{subfigure}
	\begin{subfigure}{0.3\textwidth}
		\includegraphics[width=\textwidth]{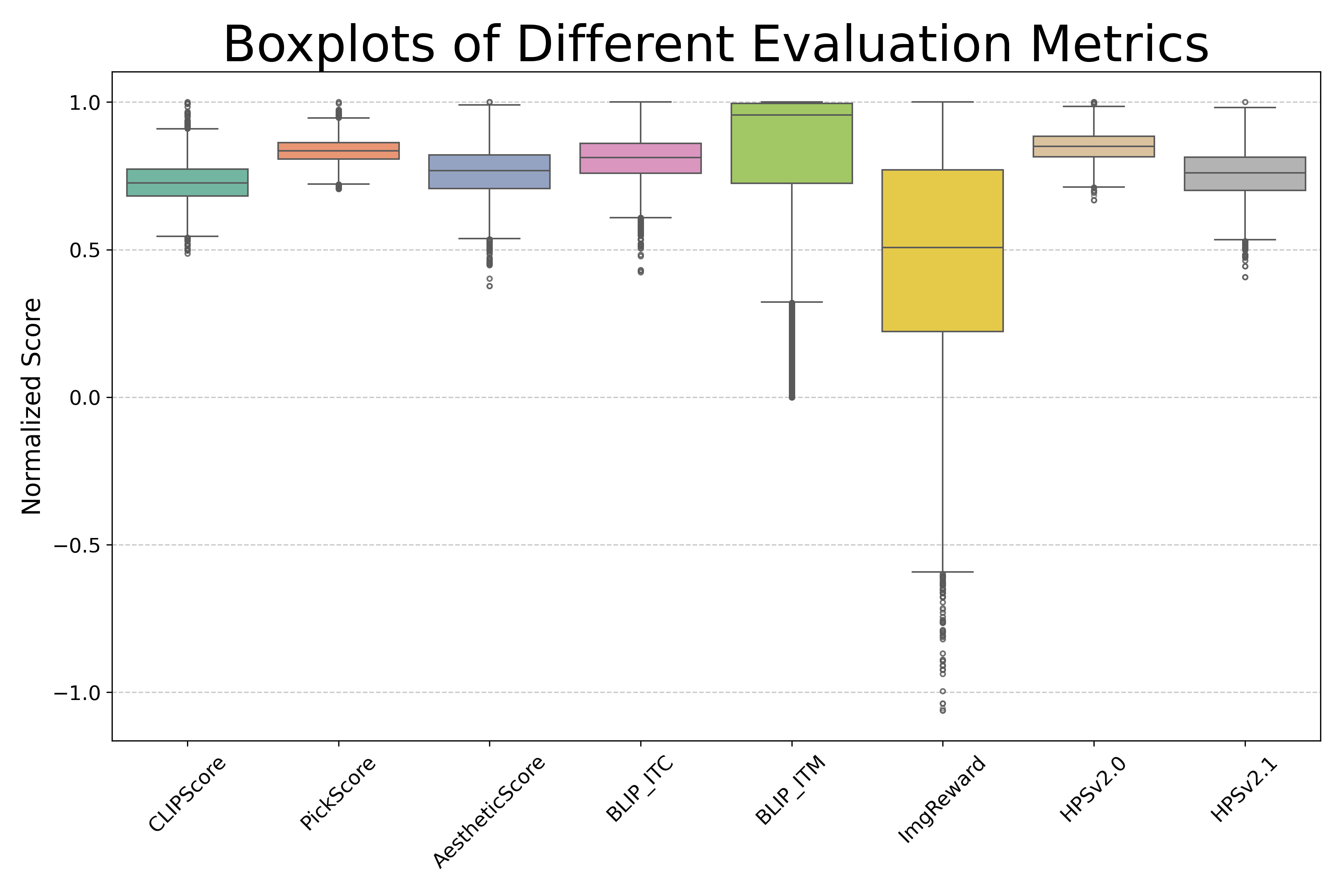}
		\caption{Flux}
	\end{subfigure}
	\caption{Boxplots of evaluation metrics of samples generated by each model. All metric scores are normalized by dividing by their respective maximum values for visualization consistency. }
	\label{fig: box plots2}
\end{figure*}

\begin{figure*}[ht]
	\centering
	\includegraphics[width=0.24\textwidth]{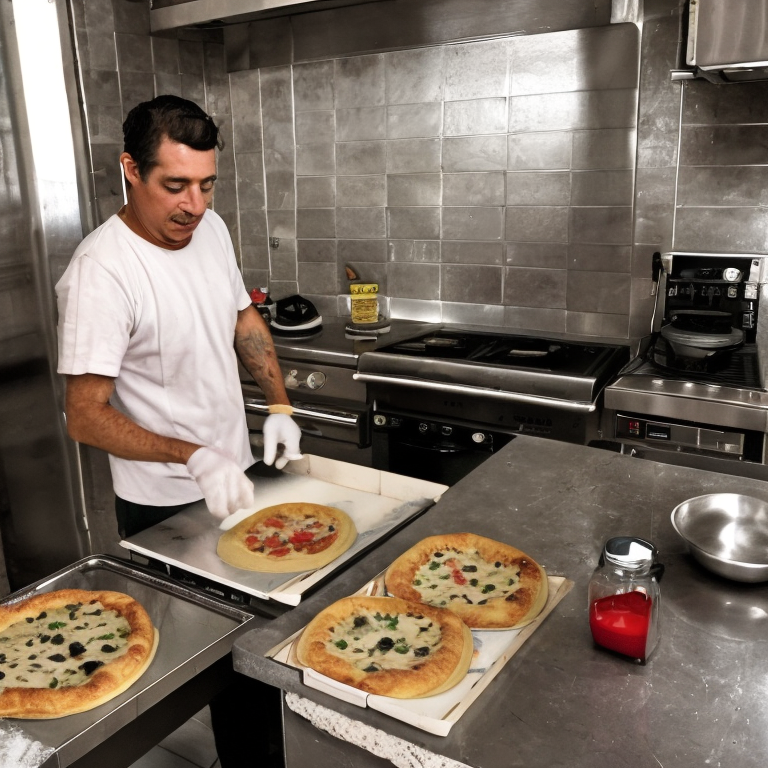}
	\includegraphics[width=0.24\textwidth]{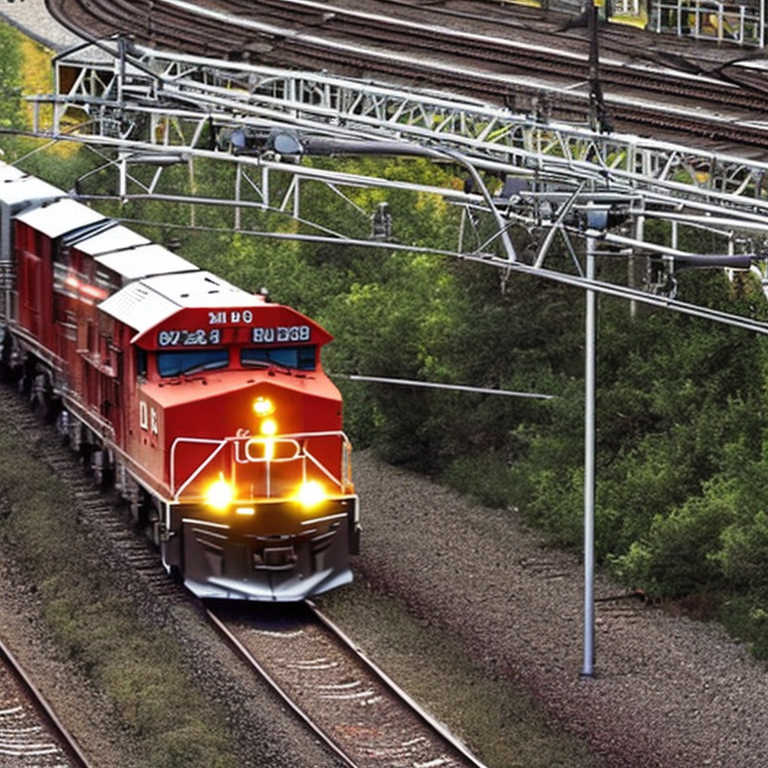}
	\includegraphics[width=0.24\textwidth]{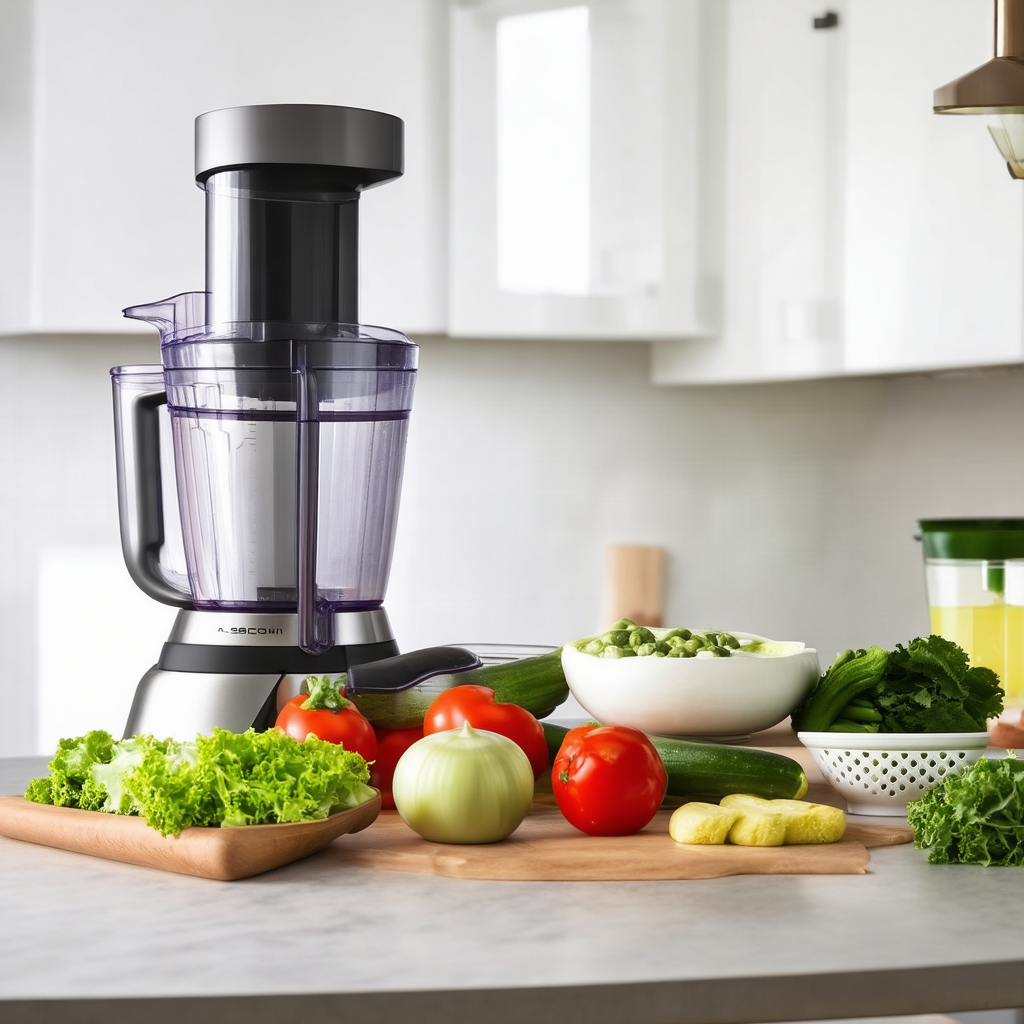}
	\includegraphics[width=0.24\textwidth]{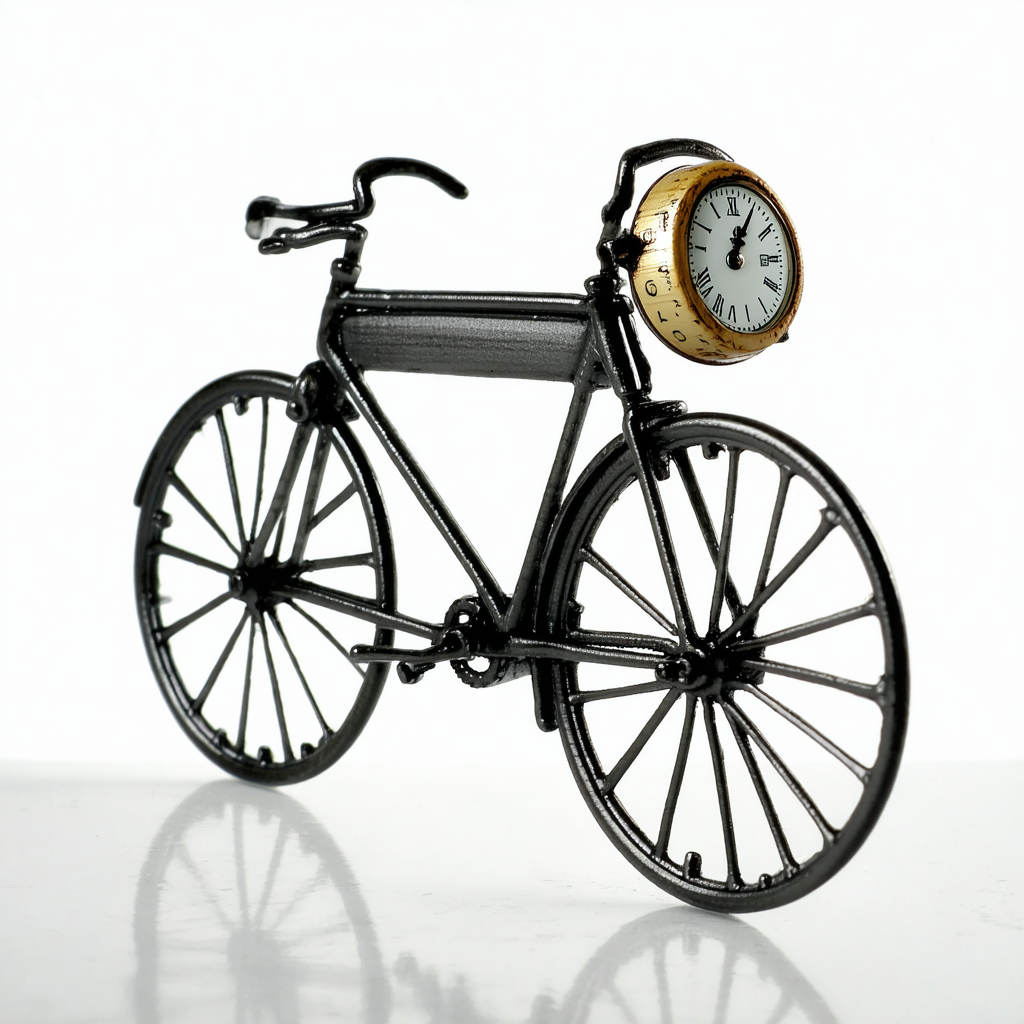}
	\caption{Final image of trajectory visualization.}
	\label{fig: final images}
\end{figure*}

\begin{figure*}[!ht]
	\centering
	% \footnotesize
	\newcommand{\tabincell}[2]{\begin{tabular}{@{}#1@{}}#2\end{tabular}}
	\begin{tabular}{m{0.6cm}<{\centering} m{1.6cm}<{\centering} m{1.6cm}<{\centering} m{1.6cm}<{\centering} m{1.6cm}<{\centering} m{1.6cm}<{\centering} m{1.6cm}<{\centering} m{1.6cm}<{\centering} m{1.6cm}<{\centering}}
		\tabincell{l}{Time}
		& \tabincell{c}{0.2}
		& \tabincell{c}{0.3}
		& \tabincell{c}{0.4}
		& \tabincell{c}{0.5}
		& \tabincell{c}{0.6}
		& \tabincell{c}{0.7}
		& \tabincell{c}{0.8}
		& \tabincell{c}{0.9}\\
		\rotatebox{90}{Down 1}
		& \includegraphics[width=18mm, height=18mm]{figs/subfig/SD2_cap1_down0_t20.png}
		& \includegraphics[width=18mm, height=18mm]{figs/subfig/SD2_cap1_down0_t30.png}
		& \includegraphics[width=18mm, height=18mm]{figs/subfig/SD2_cap1_down0_t40.png}
		& \includegraphics[width=18mm, height=18mm]{figs/subfig/SD2_cap1_down0_t50.png}
		& \includegraphics[width=18mm, height=18mm]{figs/subfig/SD2_cap1_down0_t60.png}
		& \includegraphics[width=18mm, height=18mm]{figs/subfig/SD2_cap1_down0_t70.png}
		& \includegraphics[width=18mm, height=18mm]{figs/subfig/SD2_cap1_down0_t80.png}
		& \includegraphics[width=18mm, height=18mm]{figs/subfig/SD2_cap1_down0_t90.png} \\
		\rotatebox{90}{Down 2}
		& \includegraphics[width=18mm, height=18mm]{figs/subfig/SD2_cap1_down1_t20.png}
		& \includegraphics[width=18mm, height=18mm]{figs/subfig/SD2_cap1_down1_t30.png}
		& \includegraphics[width=18mm, height=18mm]{figs/subfig/SD2_cap1_down1_t40.png}
		& \includegraphics[width=18mm, height=18mm]{figs/subfig/SD2_cap1_down1_t50.png}
		& \includegraphics[width=18mm, height=18mm]{figs/subfig/SD2_cap1_down1_t60.png}
		& \includegraphics[width=18mm, height=18mm]{figs/subfig/SD2_cap1_down1_t70.png}
		& \includegraphics[width=18mm, height=18mm]{figs/subfig/SD2_cap1_down1_t80.png}
		& \includegraphics[width=18mm, height=18mm]{figs/subfig/SD2_cap1_down1_t90.png} \\
		\rotatebox{90}{Down 3}
		& \includegraphics[width=18mm, height=18mm]{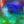}
		& \includegraphics[width=18mm, height=18mm]{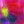}
		& \includegraphics[width=18mm, height=18mm]{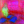}
		& \includegraphics[width=18mm, height=18mm]{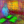}
		& \includegraphics[width=18mm, height=18mm]{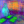}
		& \includegraphics[width=18mm, height=18mm]{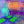}
		& \includegraphics[width=18mm, height=18mm]{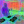}
		& \includegraphics[width=18mm, height=18mm]{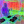} \\
		\rotatebox{90}{Down 4}
		& \includegraphics[width=18mm, height=18mm]{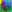}
		& \includegraphics[width=18mm, height=18mm]{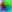}
		& \includegraphics[width=18mm, height=18mm]{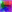}
		& \includegraphics[width=18mm, height=18mm]{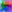}
		& \includegraphics[width=18mm, height=18mm]{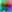}
		& \includegraphics[width=18mm, height=18mm]{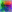}
		& \includegraphics[width=18mm, height=18mm]{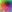}
		& \includegraphics[width=18mm, height=18mm]{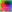} \\
		\rotatebox{90}{Mid}
		& \includegraphics[width=18mm, height=18mm]{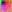}
		& \includegraphics[width=18mm, height=18mm]{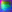}
		& \includegraphics[width=18mm, height=18mm]{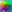}
		& \includegraphics[width=18mm, height=18mm]{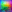}
		& \includegraphics[width=18mm, height=18mm]{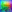}
		& \includegraphics[width=18mm, height=18mm]{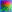}
		& \includegraphics[width=18mm, height=18mm]{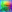}
		& \includegraphics[width=18mm, height=18mm]{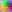} \\
		\rotatebox{90}{Up 1}
		& \includegraphics[width=18mm, height=18mm]{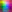}
		& \includegraphics[width=18mm, height=18mm]{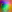}
		& \includegraphics[width=18mm, height=18mm]{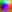}
		& \includegraphics[width=18mm, height=18mm]{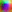}
		& \includegraphics[width=18mm, height=18mm]{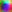}
		& \includegraphics[width=18mm, height=18mm]{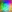}
		& \includegraphics[width=18mm, height=18mm]{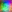}
		& \includegraphics[width=18mm, height=18mm]{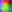} \\
		\rotatebox{90}{Up 2}
		& \includegraphics[width=18mm, height=18mm]{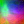}
		& \includegraphics[width=18mm, height=18mm]{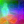}
		& \includegraphics[width=18mm, height=18mm]{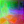}
		& \includegraphics[width=18mm, height=18mm]{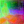}
		& \includegraphics[width=18mm, height=18mm]{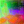}
		& \includegraphics[width=18mm, height=18mm]{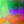}
		& \includegraphics[width=18mm, height=18mm]{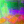}
		& \includegraphics[width=18mm, height=18mm]{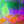} \\
		\rotatebox{90}{Up 3}
		& \includegraphics[width=18mm, height=18mm]{figs/subfig/SD2_cap1_up2_t20.png}
		& \includegraphics[width=18mm, height=18mm]{figs/subfig/SD2_cap1_up2_t30.png}
		& \includegraphics[width=18mm, height=18mm]{figs/subfig/SD2_cap1_up2_t40.png}
		& \includegraphics[width=18mm, height=18mm]{figs/subfig/SD2_cap1_up2_t50.png}
		& \includegraphics[width=18mm, height=18mm]{figs/subfig/SD2_cap1_up2_t60.png}
		& \includegraphics[width=18mm, height=18mm]{figs/subfig/SD2_cap1_up2_t70.png}
		& \includegraphics[width=18mm, height=18mm]{figs/subfig/SD2_cap1_up2_t80.png}
		& \includegraphics[width=18mm, height=18mm]{figs/subfig/SD2_cap1_up2_t90.png} \\
		\rotatebox{90}{Up 4}
		& \includegraphics[width=18mm, height=18mm]{figs/subfig/SD2_cap1_up3_t20.png}
		& \includegraphics[width=18mm, height=18mm]{figs/subfig/SD2_cap1_up3_t30.png}
		& \includegraphics[width=18mm, height=18mm]{figs/subfig/SD2_cap1_up3_t40.png}
		& \includegraphics[width=18mm, height=18mm]{figs/subfig/SD2_cap1_up3_t50.png}
		& \includegraphics[width=18mm, height=18mm]{figs/subfig/SD2_cap1_up3_t60.png}
		& \includegraphics[width=18mm, height=18mm]{figs/subfig/SD2_cap1_up3_t70.png}
		& \includegraphics[width=18mm, height=18mm]{figs/subfig/SD2_cap1_up3_t80.png}
		& \includegraphics[width=18mm, height=18mm]{figs/subfig/SD2_cap1_up3_t90.png} \\
	\end{tabular}
	\caption{PCA visualization for denoising network of Stable Diffusion 2 across time. }
	\label{fig: traj visual SD2 full}
\end{figure*}

\begin{figure*}[!ht]
	\centering
	% \footnotesize
	\newcommand{\tabincell}[2]{\begin{tabular}{@{}#1@{}}#2\end{tabular}}
	\begin{tabular}{m{0.6cm}<{\centering} m{1.6cm}<{\centering} m{1.6cm}<{\centering} m{1.6cm}<{\centering} m{1.6cm}<{\centering} m{1.6cm}<{\centering} m{1.6cm}<{\centering} m{1.6cm}<{\centering} m{1.6cm}<{\centering}}
		\tabincell{l}{Time}
		& \tabincell{c}{0.2}
		& \tabincell{c}{0.3}
		& \tabincell{c}{0.4}
		& \tabincell{c}{0.5}
		& \tabincell{c}{0.6}
		& \tabincell{c}{0.7}
		& \tabincell{c}{0.8}
		& \tabincell{c}{0.9}\\
		\rotatebox{90}{Down 1}
		& \includegraphics[width=18mm, height=18mm]{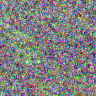}
		& \includegraphics[width=18mm, height=18mm]{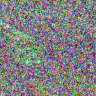}
		& \includegraphics[width=18mm, height=18mm]{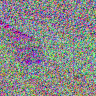}
		& \includegraphics[width=18mm, height=18mm]{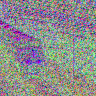}
		& \includegraphics[width=18mm, height=18mm]{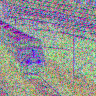}
		& \includegraphics[width=18mm, height=18mm]{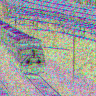}
		& \includegraphics[width=18mm, height=18mm]{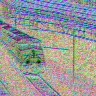}
		& \includegraphics[width=18mm, height=18mm]{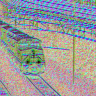} \\
		\rotatebox{90}{Down 2}
		& \includegraphics[width=18mm, height=18mm]{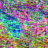}
		& \includegraphics[width=18mm, height=18mm]{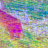}
		& \includegraphics[width=18mm, height=18mm]{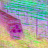}
		& \includegraphics[width=18mm, height=18mm]{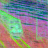}
		& \includegraphics[width=18mm, height=18mm]{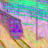}
		& \includegraphics[width=18mm, height=18mm]{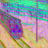}
		& \includegraphics[width=18mm, height=18mm]{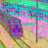}
		& \includegraphics[width=18mm, height=18mm]{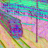} \\
		\rotatebox{90}{Down 3}
		& \includegraphics[width=18mm, height=18mm]{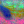}
		& \includegraphics[width=18mm, height=18mm]{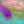}
		& \includegraphics[width=18mm, height=18mm]{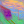}
		& \includegraphics[width=18mm, height=18mm]{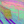}
		& \includegraphics[width=18mm, height=18mm]{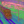}
		& \includegraphics[width=18mm, height=18mm]{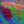}
		& \includegraphics[width=18mm, height=18mm]{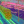}
		& \includegraphics[width=18mm, height=18mm]{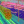} \\
		\rotatebox{90}{Down 4}
		& \includegraphics[width=18mm, height=18mm]{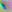}
		& \includegraphics[width=18mm, height=18mm]{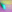}
		& \includegraphics[width=18mm, height=18mm]{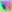}
		& \includegraphics[width=18mm, height=18mm]{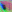}
		& \includegraphics[width=18mm, height=18mm]{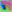}
		& \includegraphics[width=18mm, height=18mm]{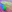}
		& \includegraphics[width=18mm, height=18mm]{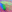}
		& \includegraphics[width=18mm, height=18mm]{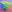} \\
		\rotatebox{90}{Mid}
		& \includegraphics[width=18mm, height=18mm]{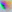}
		& \includegraphics[width=18mm, height=18mm]{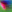}
		& \includegraphics[width=18mm, height=18mm]{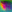}
		& \includegraphics[width=18mm, height=18mm]{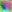}
		& \includegraphics[width=18mm, height=18mm]{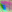}
		& \includegraphics[width=18mm, height=18mm]{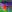}
		& \includegraphics[width=18mm, height=18mm]{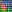}
		& \includegraphics[width=18mm, height=18mm]{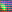} \\
		\rotatebox{90}{Up 1}
		& \includegraphics[width=18mm, height=18mm]{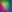}
		& \includegraphics[width=18mm, height=18mm]{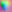}
		& \includegraphics[width=18mm, height=18mm]{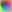}
		& \includegraphics[width=18mm, height=18mm]{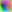}
		& \includegraphics[width=18mm, height=18mm]{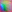}
		& \includegraphics[width=18mm, height=18mm]{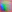}
		& \includegraphics[width=18mm, height=18mm]{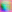}
		& \includegraphics[width=18mm, height=18mm]{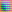} \\
		\rotatebox{90}{Up 2}
		& \includegraphics[width=18mm, height=18mm]{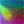}
		& \includegraphics[width=18mm, height=18mm]{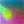}
		& \includegraphics[width=18mm, height=18mm]{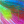}
		& \includegraphics[width=18mm, height=18mm]{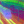}
		& \includegraphics[width=18mm, height=18mm]{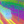}
		& \includegraphics[width=18mm, height=18mm]{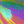}
		& \includegraphics[width=18mm, height=18mm]{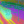}
		& \includegraphics[width=18mm, height=18mm]{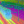} \\
		\rotatebox{90}{Up 3}
		& \includegraphics[width=18mm, height=18mm]{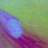}
		& \includegraphics[width=18mm, height=18mm]{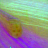}
		& \includegraphics[width=18mm, height=18mm]{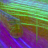}
		& \includegraphics[width=18mm, height=18mm]{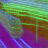}
		& \includegraphics[width=18mm, height=18mm]{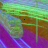}
		& \includegraphics[width=18mm, height=18mm]{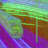}
		& \includegraphics[width=18mm, height=18mm]{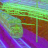}
		& \includegraphics[width=18mm, height=18mm]{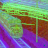} \\
		\rotatebox{90}{Up 4}
		& \includegraphics[width=18mm, height=18mm]{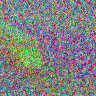}
		& \includegraphics[width=18mm, height=18mm]{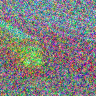}
		& \includegraphics[width=18mm, height=18mm]{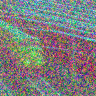}
		& \includegraphics[width=18mm, height=18mm]{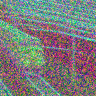}
		& \includegraphics[width=18mm, height=18mm]{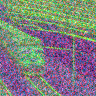}
		& \includegraphics[width=18mm, height=18mm]{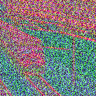}
		& \includegraphics[width=18mm, height=18mm]{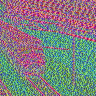}
		& \includegraphics[width=18mm, height=18mm]{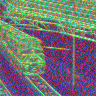} \\
	\end{tabular}
	\caption{PCA visualization for denoising network of Stable Diffusion 2 across time. }
	\label{fig: traj visual SD2 ADD}
\end{figure*}

\begin{figure*}[!ht]
	\centering
	% \footnotesize
	\newcommand{\tabincell}[2]{\begin{tabular}{@{}#1@{}}#2\end{tabular}}
	\begin{tabular}{m{0.6cm}<{\centering} m{1.6cm}<{\centering} m{1.6cm}<{\centering} m{1.6cm}<{\centering} m{1.6cm}<{\centering} m{1.6cm}<{\centering} m{1.6cm}<{\centering} m{1.6cm}<{\centering} m{1.6cm}<{\centering}}
		\tabincell{l}{Time}
		& \tabincell{c}{0.2}
		& \tabincell{c}{0.3}
		& \tabincell{c}{0.4}
		& \tabincell{c}{0.5}
		& \tabincell{c}{0.6}
		& \tabincell{c}{0.7}
		& \tabincell{c}{0.8}
		& \tabincell{c}{0.9}\\
		\rotatebox{90}{Block 0}
		& \includegraphics[width=18mm, height=18mm]{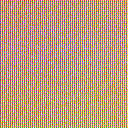}
		& \includegraphics[width=18mm, height=18mm]{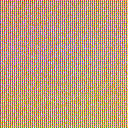}
		& \includegraphics[width=18mm, height=18mm]{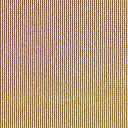}
		& \includegraphics[width=18mm, height=18mm]{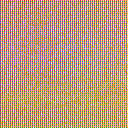}
		& \includegraphics[width=18mm, height=18mm]{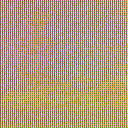}
		& \includegraphics[width=18mm, height=18mm]{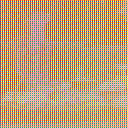}
		& \includegraphics[width=18mm, height=18mm]{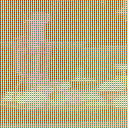}
		& \includegraphics[width=18mm, height=18mm]{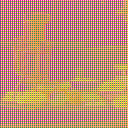} \\
		\rotatebox{90}{Block 4}
		& \includegraphics[width=18mm, height=18mm]{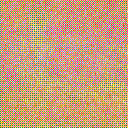}
		& \includegraphics[width=18mm, height=18mm]{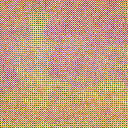}
		& \includegraphics[width=18mm, height=18mm]{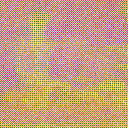}
		& \includegraphics[width=18mm, height=18mm]{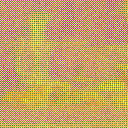}
		& \includegraphics[width=18mm, height=18mm]{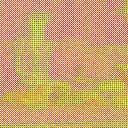}
		& \includegraphics[width=18mm, height=18mm]{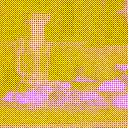}
		& \includegraphics[width=18mm, height=18mm]{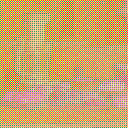}
		& \includegraphics[width=18mm, height=18mm]{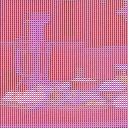} \\
		\rotatebox{90}{Block 8}
		& \includegraphics[width=18mm, height=18mm]{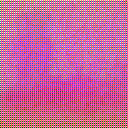}
		& \includegraphics[width=18mm, height=18mm]{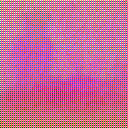}
		& \includegraphics[width=18mm, height=18mm]{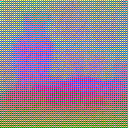}
		& \includegraphics[width=18mm, height=18mm]{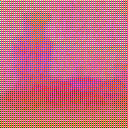}
		& \includegraphics[width=18mm, height=18mm]{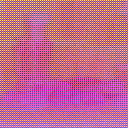}
		& \includegraphics[width=18mm, height=18mm]{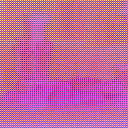}
		& \includegraphics[width=18mm, height=18mm]{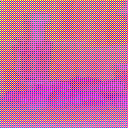}
		& \includegraphics[width=18mm, height=18mm]{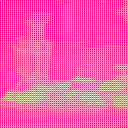} \\
		\rotatebox{90}{Block 12}
		& \includegraphics[width=18mm, height=18mm]{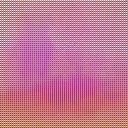}
		& \includegraphics[width=18mm, height=18mm]{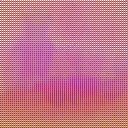}
		& \includegraphics[width=18mm, height=18mm]{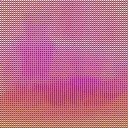}
		& \includegraphics[width=18mm, height=18mm]{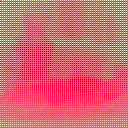}
		& \includegraphics[width=18mm, height=18mm]{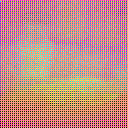}
		& \includegraphics[width=18mm, height=18mm]{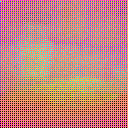}
		& \includegraphics[width=18mm, height=18mm]{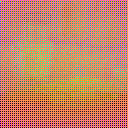}
		& \includegraphics[width=18mm, height=18mm]{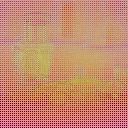} \\
		\rotatebox{90}{Block 16}
		& \includegraphics[width=18mm, height=18mm]{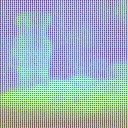}
		& \includegraphics[width=18mm, height=18mm]{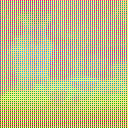}
		& \includegraphics[width=18mm, height=18mm]{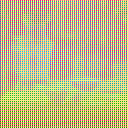}
		& \includegraphics[width=18mm, height=18mm]{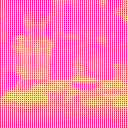}
		& \includegraphics[width=18mm, height=18mm]{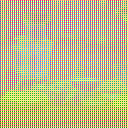}
		& \includegraphics[width=18mm, height=18mm]{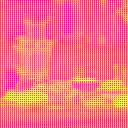}
		& \includegraphics[width=18mm, height=18mm]{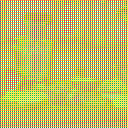}
		& \includegraphics[width=18mm, height=18mm]{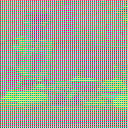} \\
		\rotatebox{90}{Block 20}
		& \includegraphics[width=18mm, height=18mm]{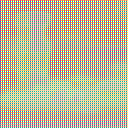}
		& \includegraphics[width=18mm, height=18mm]{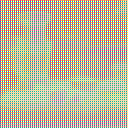}
		& \includegraphics[width=18mm, height=18mm]{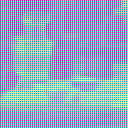}
		& \includegraphics[width=18mm, height=18mm]{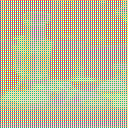}
		& \includegraphics[width=18mm, height=18mm]{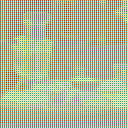}
		& \includegraphics[width=18mm, height=18mm]{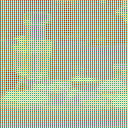}
		& \includegraphics[width=18mm, height=18mm]{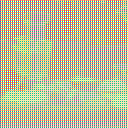}
		& \includegraphics[width=18mm, height=18mm]{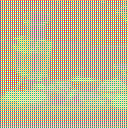} \\
	\end{tabular}
	\caption{PCA visualization for denoising network of Stable Diffusion 3.5 medium across time. }
	\label{fig: traj visual SD3M ADD}
\end{figure*}

\begin{figure*}[!ht]
	\centering
	% \footnotesize
	\newcommand{\tabincell}[2]{\begin{tabular}{@{}#1@{}}#2\end{tabular}}
	\begin{tabular}{m{0.6cm}<{\centering} m{1.6cm}<{\centering} m{1.6cm}<{\centering} m{1.6cm}<{\centering} m{1.6cm}<{\centering} m{1.6cm}<{\centering} m{1.6cm}<{\centering} m{1.6cm}<{\centering} m{1.6cm}<{\centering}}
		\tabincell{l}{Time}
		& \tabincell{c}{0.2}
		& \tabincell{c}{0.3}
		& \tabincell{c}{0.4}
		& \tabincell{c}{0.5}
		& \tabincell{c}{0.6}
		& \tabincell{c}{0.7}
		& \tabincell{c}{0.8}
		& \tabincell{c}{0.9}\\
		\rotatebox{90}{Block 0}
		& \includegraphics[width=18mm, height=18mm]{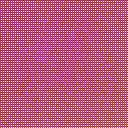}
		& \includegraphics[width=18mm, height=18mm]{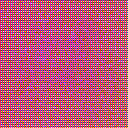}
		& \includegraphics[width=18mm, height=18mm]{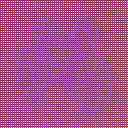}
		& \includegraphics[width=18mm, height=18mm]{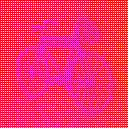}
		& \includegraphics[width=18mm, height=18mm]{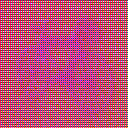}
		& \includegraphics[width=18mm, height=18mm]{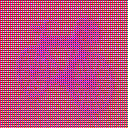}
		& \includegraphics[width=18mm, height=18mm]{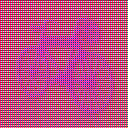}
		& \includegraphics[width=18mm, height=18mm]{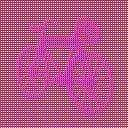} \\
		\rotatebox{90}{Block 4}
		& \includegraphics[width=18mm, height=18mm]{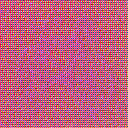}
		& \includegraphics[width=18mm, height=18mm]{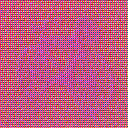}
		& \includegraphics[width=18mm, height=18mm]{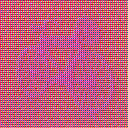}
		& \includegraphics[width=18mm, height=18mm]{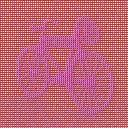}
		& \includegraphics[width=18mm, height=18mm]{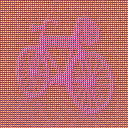}
		& \includegraphics[width=18mm, height=18mm]{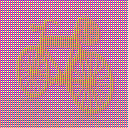}
		& \includegraphics[width=18mm, height=18mm]{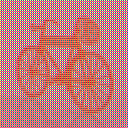}
		& \includegraphics[width=18mm, height=18mm]{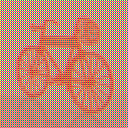} \\
		\rotatebox{90}{Block 8}
		& \includegraphics[width=18mm, height=18mm]{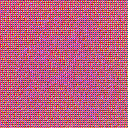}
		& \includegraphics[width=18mm, height=18mm]{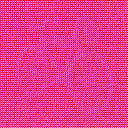}
		& \includegraphics[width=18mm, height=18mm]{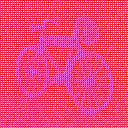}
		& \includegraphics[width=18mm, height=18mm]{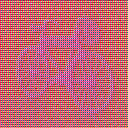}
		& \includegraphics[width=18mm, height=18mm]{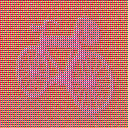}
		& \includegraphics[width=18mm, height=18mm]{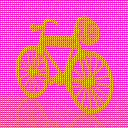}
		& \includegraphics[width=18mm, height=18mm]{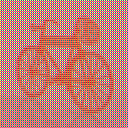}
		& \includegraphics[width=18mm, height=18mm]{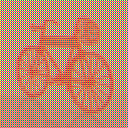} \\
		\rotatebox{90}{Block 12}
		& \includegraphics[width=18mm, height=18mm]{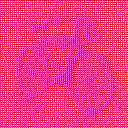}
		& \includegraphics[width=18mm, height=18mm]{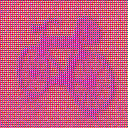}
		& \includegraphics[width=18mm, height=18mm]{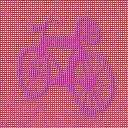}
		& \includegraphics[width=18mm, height=18mm]{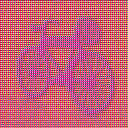}
		& \includegraphics[width=18mm, height=18mm]{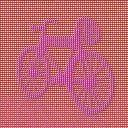}
		& \includegraphics[width=18mm, height=18mm]{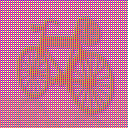}
		& \includegraphics[width=18mm, height=18mm]{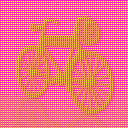}
		& \includegraphics[width=18mm, height=18mm]{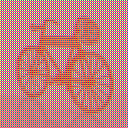} \\
		\rotatebox{90}{Block 16}
		& \includegraphics[width=18mm, height=18mm]{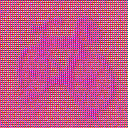}
		& \includegraphics[width=18mm, height=18mm]{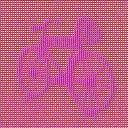}
		& \includegraphics[width=18mm, height=18mm]{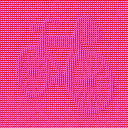}
		& \includegraphics[width=18mm, height=18mm]{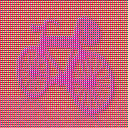}
		& \includegraphics[width=18mm, height=18mm]{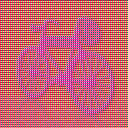}
		& \includegraphics[width=18mm, height=18mm]{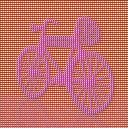}
		& \includegraphics[width=18mm, height=18mm]{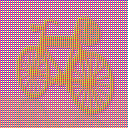}
		& \includegraphics[width=18mm, height=18mm]{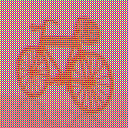} \\
		\rotatebox{90}{Block 20}
		& \includegraphics[width=18mm, height=18mm]{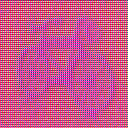}
		& \includegraphics[width=18mm, height=18mm]{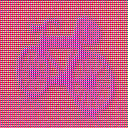}
		& \includegraphics[width=18mm, height=18mm]{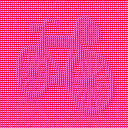}
		& \includegraphics[width=18mm, height=18mm]{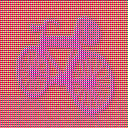}
		& \includegraphics[width=18mm, height=18mm]{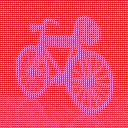}
		& \includegraphics[width=18mm, height=18mm]{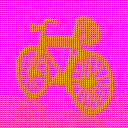}
		& \includegraphics[width=18mm, height=18mm]{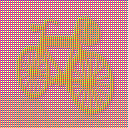}
		& \includegraphics[width=18mm, height=18mm]{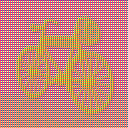} \\
		\rotatebox{90}{Block 24}
		& \includegraphics[width=18mm, height=18mm]{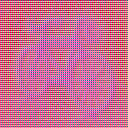}
		& \includegraphics[width=18mm, height=18mm]{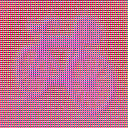}
		& \includegraphics[width=18mm, height=18mm]{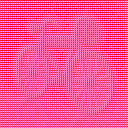}
		& \includegraphics[width=18mm, height=18mm]{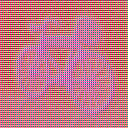}
		& \includegraphics[width=18mm, height=18mm]{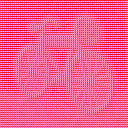}
		& \includegraphics[width=18mm, height=18mm]{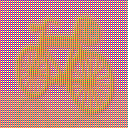}
		& \includegraphics[width=18mm, height=18mm]{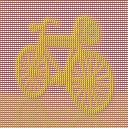}
		& \includegraphics[width=18mm, height=18mm]{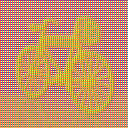} \\
		\rotatebox{90}{Block 28}
		& \includegraphics[width=18mm, height=18mm]{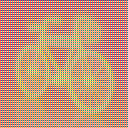}
		& \includegraphics[width=18mm, height=18mm]{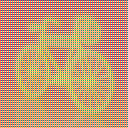}
		& \includegraphics[width=18mm, height=18mm]{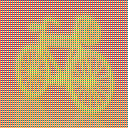}
		& \includegraphics[width=18mm, height=18mm]{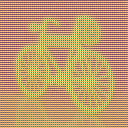}
		& \includegraphics[width=18mm, height=18mm]{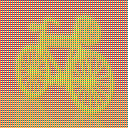}
		& \includegraphics[width=18mm, height=18mm]{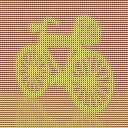}
		& \includegraphics[width=18mm, height=18mm]{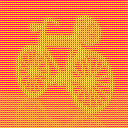}
		& \includegraphics[width=18mm, height=18mm]{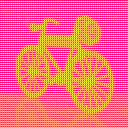} \\
		\rotatebox{90}{Block 32}
		& \includegraphics[width=18mm, height=18mm]{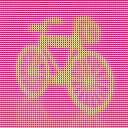}
		& \includegraphics[width=18mm, height=18mm]{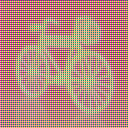}
		& \includegraphics[width=18mm, height=18mm]{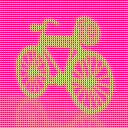}
		& \includegraphics[width=18mm, height=18mm]{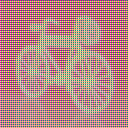}
		& \includegraphics[width=18mm, height=18mm]{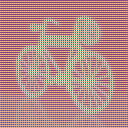}
		& \includegraphics[width=18mm, height=18mm]{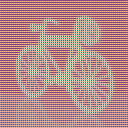}
		& \includegraphics[width=18mm, height=18mm]{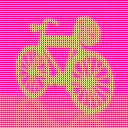}
		& \includegraphics[width=18mm, height=18mm]{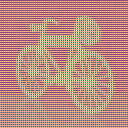} \\
	\end{tabular}
	\caption{PCA visualization for denoising network of Stable Diffusion 3.5 large across time. }
	\label{fig: traj visual SD3L ADD}
\end{figure*}

\section{Implementation Details }\label{app:Probe-Select-impl}
\subsection{Network Structure and Data Processing Pipeline}
\noindent\textbf{Signals and Inputs.} During sampling, we tap an intermediate denoiser activation $h_t \in \mathbb{R}^{C\times H\times W}$ at an early timestep $t$ and optionally a text-side embedding $e_{\text{text}}$ depending on the target evaluator. A sinusoidal timestep embedding is computed and projected to $e_t \in \mathbb{R}^{d_t}$, which conditions subsequent residual blocks. In addition, for SD3 we first un-patch $h_t$ to recover the dense feature map (this step is not needed for SD2). We then resize the feature map by bilinear interpolation to $C \times 48 \times 48$. Next, we apply PCA along the channel dimension and keep the top $48$ components, which yields a tensor of size $48 \times 48 \times 48$. This compression reduces GPU memory by a large margin and, in our experiments, does not cause a noticeable drop in evaluation performance. 

\noindent\textbf{Time-Conditioned Feature Encoder.}
We encode $h_t$ using a stack of attention--residual down blocks with time modulation. Each block contains a time-conditioned ResNet sub-block followed by self-attention, and $3\times3$ convolution downsampling. 

We use GroupNorm with $16$ groups in the residual blocks, \texttt{SiLU} as the residual activation, and \texttt{conv} downsampling. The attention head dimension is set to $5$. The sinusoidal timestep embedding has $64$ channels and is projected to a time MLP of width $512$; the residual blocks consume this time embedding through scale–shift conditioning. 

The resulting feature map is aggregated by average pooling and flattened to a compact vector, which is then passed through a two-layer image-side MLP with one hidden layer of width $512$ (\texttt{ReLU}, dropout $0.5$, BatchNorm enabled). 

\noindent\textbf{Optional Text Alignment.}
For evaluators that depend on the prompt semantics (e.g., CLIP- or BLIP-based metrics, ImageReward, HPS), we project the text embedding to the same 512-dimensional space. We then form a concatenated vector by stacking $u$, the text vector $v$, their element-wise product $u\odot v$, and the absolute difference $|u-v|$. This enriched representation is fed to the final prediction head. We also compute a cosine-similarity matrix between $u$ and $v$ for analysis. 

\noindent\textbf{Prediction Head.}
A three-layer MLP (hidden width $512, 256$) with a Sigmoid output maps the representation to a scalar in $[0,1]$ that serves as the early quality score. 

\noindent\textbf{Head Multiplicity and Dimensions.}
We instantiate one probe network per target evaluator key (e.g., \texttt{CLIPScore}, \texttt{BLIP\_ITM}, \texttt{ImgReward}, \texttt{HPSv21}). Keys that use text rely on a text projector, while text-free keys skip this branch. Implementation uses $d_t{=}64$ for the sinusoidal timestep input, a time-MLP width of $512$, attention groups of $32$, and a final two-layer MLP with hidden sizes $[512, 256]$ and Sigmoid output. Pooling uses a $3{\times}3$ average operator by default.

\section{Additional Results}\label{appendix: additional results}
\subsection{Full Visualization}
This section provides comprehensive visualizations of the structural signals evolving within the denoiser network across different timesteps, supplementing the core finding in Sec. \ref{sec: Early Structural Evidence in Denoiser Features}. We visualize the feature maps by applying PCA along the channel dimension for a better view of their underlying structure, similar to Fig. \ref{fig: traj visual} in main text.

\textbf{Trajectory Final Images}. Figure \ref{fig: final images} shows the final generated images for the trajectories visualized in this section. 

\textbf{SD2 Full PCA Visualization}. Fig. \ref{fig: traj visual SD2 full} and Fig. \ref{fig: traj visual SD2 ADD} extend the analysis of Stable Diffusion 2 (SD2) features. They show the PCA visualization for all Down, Mid, and Up blocks across timesteps $t=0.2$ to $t=0.9$ for two different prompts. This full set confirms the observation: while high-resolution details are absent early on, coarse structural cues (like layout and large object boundaries) emerge and stabilize in the mid-to-late layers (especially Up $\mathbf{3}$ as selected for probing) as early as $t=0.2$ to $t=0.3$. The features in Up $\mathbf{3}$ remain relatively stable compared to the noisier, high-frequency layers (Down 1, Up 4). 

\textbf{SD3-M \& SD3-L Full PCA Visualization}.
Fig. \ref{fig: traj visual SD3M ADD} and Fig. \ref{fig: traj visual SD3L ADD} present the PCA visualizations for the intermediate blocks of Stable Diffusion 3.5 Medium (SD3-M) and Stable Diffusion 3.5 Large (SD3-L), respectively. Since SD3 series model contains more than 20 blocks, we visualize the feature every 4 blocks (strat from 0). 
These visualizations confirm that the structural stability phenomenon generalizes to transformer-based denoisers as well. Layers corresponding to mid-to-high resolutions (e.g., Block 8 and Block 12 in SD3-M, Block 16 and Block 20 in SD3-L) rapidly converge to a stable representation of the main objects and their layout (e.g., the bicycle shape and its components), even when $t \leq 0.3$. This cross-backbone consistency further validates the structural signal discovery for early quality assessment. 

Finally, we use the output of block 20 of SD3-M and block 28 of SD3-L for representative feature for Probe-Select. 

\subsection{Supplement Results for Main Paper}
This section provides additional quantitative and visual results referenced in the main paper, including the full-time range of Spearman correlations (extending Tab. \ref{tab: spearm corre}) and the boxplots for all generative backbones (extending Fig. \ref{fig: box plots}). 

Table \ref{tab: full spearm corre} presents the additional set of Spearman correlations between the early probe predictions ( $\hat{y}_{t, m}$ ) and the final ground-truth metric scores ( $R_m\left(x_1\right)$ ) across four backbones and eight evaluators. This table supplements the subset of results presented in the main paper's  Tab. \ref{tab: spearm corre} by including correlations at timesteps $t=0.3$ and $t=0.5$. 

The complete results confirm the primary finding of high and stable correlations : the correlation scores across all metrics are already strong at $t=0.2$ and remain highly stable, changing only marginally up to $t=0.6$. Specifically, metrics like BLIP-ITM and ImageReward achieve correlations near $0.98-0.99$ across all checkpoints, validating that reliable ranking signals emerge very early in the diffusion process, making $t=0.2$ a robust choice for early pruning.

Figure \ref{fig: box plots2} provides the boxplots for the normalized scores of all eight evaluation metrics (CLIPScore, PickScore, AeS, etc.) on samples generated by SD2, SD3-M, and FLUX.1-dev. These plots are provided for completeness (as referenced in \ref{fig: box plots}) and illustrate the inherent distribution and dynamic range of each reference metric across the samples. 

As noted in the main text Sec. \ref{sec: Quantitative Analysis: Predicting Final Quality from Partial States}, metrics with a broader, less-saturated score distribution, such as ImageReward (ImgReward), naturally produce a more stable relative ordering, which translates into the higher Spearman correlations. Conversely, metrics with narrower distributions, such as the HPSv2.x scores, often result in tighter clusters and a greater likelihood of ties in ranking.

\subsection{Earlier Checkpoints and Scheduler Robustness}\label{app:earlyt}
To justify the default choice of $t=0.2$, we further evaluate earlier checkpoints on SD3-L. Table~\ref{tab:early_t_choice} shows that the correlation is still weak at very early stages such as $t=0.05$, improves substantially at $t=0.1$ and $t=0.15$, and becomes consistently strong by $t=0.2$. This supports our claim that $t=0.2$ offers a strong balance between prediction quality and compute savings.

We also repeat the analysis under different samplers, including Euler and Heun schedulers, and observe similar Spearman correlations at the same normalized checkpoint. This indicates that the early structural signal is robust to the specific scheduler choice and is not tied to one sampling implementation. 
\begin{table}[!ht]
	\centering
	\footnotesize
	\begin{tabular}{c|cccccc}
		\toprule
		$t$ & CS & PS & AS & BIC & BIM & IR \\
		\midrule
		0.05 & 0.24 & 0.16 & 0.18 & 0.16 & 0.28 & 0.29\\
		0.10 & 0.64 & 0.68 & 0.57 & 0.62 & 0.76 & 0.78\\
		0.15 & 0.75 & 0.82 & 0.67 & 0.69 & 0.92 & 0.92\\
		0.20 & 0.79 & 0.84 & 0.70 & 0.74 & 0.98 & 0.99\\
		\bottomrule
	\end{tabular}
	\caption{Spearman correlation at very early checkpoints on SD3-L. Correlations become consistently strong at $t=0.2$, supporting our default choice.}
	\label{tab:early_t_choice}
\end{table}

\subsection{Cross-Backbone Transfer of Probe Networks}
Although our main experiments train one probe per backbone, we find that the learned probe transfers well across different diffusion models after the shared PCA-based feature processing. Table~\ref{tab:cross_backbone_transfer} reports the transfer results across SD2, SD3-M, SD3-L, and FLUX.1-dev. A probe trained on one backbone remains close to peak performance on the others, suggesting that the captured early structural signal is not highly model-specific. This substantially reduces the practical cost of deployment, since one trained probe can serve as a plug-in evaluator for multiple backbones. 

\begin{table}[!ht]
	\centering
	\footnotesize
	\begin{tabular}{c|cccc}
		\toprule
		\diagbox{Source}{Target} & SD2 & SD3-M & SD3-L & FLUX.1-dev \\ 
		\midrule
		SD2 & 0.98 & 0.96 & 0.97 & 0.96 \\ 
		SD3-M & 0.86 & 0.98 & 0.95 & 0.98 \\ 
		SD3-L & 0.86 & 0.98 & 0.98 & 0.97 \\ 
		FLUX.1-dev & 0.89 & 0.97 & 0.97 & 0.98 \\ 
		\bottomrule
	\end{tabular}
	\caption{Cross-backbone transfer of probe networks. Each entry reports the Spearman correlation when a probe trained on the source backbone is applied to the target backbone.}
	\label{tab:cross_backbone_transfer}
\end{table}

\subsection{Additional Baseline: Decoding and Scoring Early}
A simple alternative to our method is to decode the predicted clean image $\hat{x}_0$ at an early checkpoint and directly apply a standard evaluator. We test this baseline at $t=0.2$ using BLIP-ITM (BIM). The resulting Spearman correlation is only around $0.52$, which is much lower than the correlation achieved by Probe-Select (typically above $0.9$ under the same setting).

This gap is expected: standard evaluators are trained on fully formed images and are not designed to assess the potential of blurry or partially denoised outputs. In contrast, Probe-Select is trained directly on intermediate denoiser features and learns to map early structural signals to the final quality. This result shows that early quality assessment requires dedicated probing rather than directly reusing final-image evaluators on incomplete samples.

\begin{table}[h]
	\centering
	\footnotesize
	\begin{tabular}{c|c|ccccc}
		\toprule
		Dataset & Method & CS & PS & BIC & BIM & IR \\
		\midrule
		\multirow{2}{*}{DrawBench}
		& Baseline & 0.32 & 0.23 & 0.48 & 0.90 & 1.08 \\
		& Probe-Select & 0.35 & 0.23 & 0.52 & 0.93 & 1.55 \\
		\midrule
		\multirow{2}{*}{GenEval}
		& Baseline & 0.34 & 0.23 & 0.49 & 0.90 & 1.07 \\
		& Probe-Select & 0.35 & 0.23 & 0.52 & 0.96 & 1.53 \\
		\midrule
		\multirow{2}{*}{HPD}
		& Baseline & 0.34 & 0.22 & 0.48 & 0.95 & 1.12 \\
		& Probe-Select & 0.36 & 0.23 & 0.51 & 0.98 & 1.38 \\
		\midrule
		\multirow{2}{*}{T2I-CompBench}
		& Baseline & 0.32 & 0.22 & 0.49 & 0.95 & 1.13 \\
		& Probe-Select & 0.33 & 0.23 & 0.52 & 0.98 & 1.49 \\
		\bottomrule
	\end{tabular}
	\caption{Generalization to additional text-to-image benchmarks. Probe-Select consistently improves final quality metrics over the no-selection baseline.}
	\label{tab:generalization_benchmarks}
\end{table}

\subsection{Ablation and Sensitivity Analysis}
We conduct an ablation study to understand the influence of the key hyperparameters governing our joint training objective: the listwise ranking temperature $\tau_{\text {list,max, }}$ the listwise ranking margin $\alpha_{\text {max }}$, and the weight of the contrastive alignment loss $\lambda_{\text {Align }}$. All ablations are performed on the FLUX.1-dev backbone using the ImageReward evaluator as the target metric, with probe predictions taken at the early checkpoint $t=0.2$. The baseline configuration uses the values $\tau_{\text {list, max}}=\tau_{\text{Align, max }}=1.0, \alpha_{\max}=0.4 \sigma$, and $\lambda_{\text {Align }}=10$.

\textbf{Effect of Alignment Loss Weight ( $\lambda_{\text {Align }}$ ).}
The weight $\lambda_{\text {Align }}$ balances the primary ranking objective $\mathcal{L}_{\text {list }}$ and the auxiliary semantic alignment objective $\mathcal{L}_{\text {Align }}$ (Equation 7). $\mathcal{L}_{\text {Align }}$ helps maintain prompt awareness in the probe's latent space. We investigate its values with $\tau_{\text {list,max }}=1.0$ and $\alpha_{\text {max }}=0.4 \sigma$ fixed. The $\lambda_{\text {Align }}=0$ case corresponds to only using the listwise ranking loss.

Results presented in Table 6 show that the contrastive alignment loss is crucial for high-fidelity quality prediction. Without it ( $\lambda_{\text {Align }}=0$ ), the Spearman correlation drops significantly to 0.66 . The correlation improves steadily as $\lambda_{\text {Align }}$ increases, reaching its peak fidelity of 0.99 at the baseline value of $\lambda_{\text {Align }}=10$. This demonstrates that simply learning to rank based on visual cues is insufficient; the probe must also be guided to align its representation with the text prompt embedding to successfully forecast text-aware quality metrics like ImageReward. The slight drop at $\lambda_{\text {Align }}=50$ suggests an overly strong focus on text alignment can hurt ranking performance.

\begin{table}[!ht]
	\centering
	\begin{tabular}{cc}
		\toprule
		$\tau$ & Validation Spearman Correlation (IR) \\
		\midrule
		0 &  0.66\\
		1 &  0.79\\
		2 &  0.84\\
		5 &  0.94\\
		10 & 0.99\\
		20 & 0.98\\
		50 & 0.93\\
		\bottomrule
	\end{tabular}
	\caption{Ablation on the contrastive alignment loss weight $\lambda_\text{Align}$}
	\label{tab: ab3}
\end{table}

\textbf{Effect of Temperature ( $\tau_{\text{list, max}}$ and $\tau_{\text{Align, max }}$).}
The temperature $\tau$ controls the smoothness of the softmax functions in both the listwise ranking loss ( $\mathcal{L}_{\text {list }}$ ) and the contrastive loss ( $\mathcal{L}_{\text {Align }}$ ), where we set $\tau_{\text {list,max }}=\tau_{\text {Align,max }}$. A lower temperature leads to a stricter loss. We investigate its effect while keeping $\alpha_{\max }=0.4 \sigma$ and $\lambda_{\text {Align }}=10$ fixed.

As shown in Tab. \ref{tab: ab1}, the Spearman correlation is highly robust to the choice of $\tau$ in the range of $[0.1,10.0]$, remaining exceptionally high across all tested values (between 0.96 and 0.99 ). The best performance is achieved with $\tau \in\{0.5,1.0\}$, indicating that a moderate degree of strictness in the ranking is beneficial. The general stability suggests that once the prompt-aware features are extracted (due to $\lambda_{\text {Align }}>0$ ), the precise shaping of the ranking loss via the temperature is less critical.

\begin{table}[!ht]
	\centering
	\begin{tabular}{cc}
		\toprule
		$\tau$ & Validation Spearman Correlation (IR) \\
		\midrule
		0.1 & 0.98 \\
		0.5 & 0.99 \\
		1.0 & 0.99 \\
		2.0 & 0.97 \\
		3.0 & 0.96 \\
		5.0 & 0.96 \\
		10.0 & 0.96 \\
		\bottomrule
	\end{tabular}
	\caption{Table 4. Ablation on the temperature $\tau_{\text{list, max}}$ and $\tau_{\text{Align, max }}$. }
	\label{tab: ab1}
\end{table}

\textbf{Effect of Listwise Ranking Margin ( $\alpha_{\max}$ ).}
The margin $\alpha_{\text {max }}$ in $\mathcal{L}_{\text {list }}$ ensures a minimum score separation is enforced between preferred and less-preferred samples. We study its impact with $\tau_{\text {list,max }}=1.0$ and $\lambda_{\text {Align }}=10$ fixed. The margin is expressed as a fraction of $\sigma$, the standard deviation of the ImageReward scores.

The results in Table \ref{tab: ab2} indicate that the margin is a critical hyperparameter. Very small margins, such as $0.01 \sigma$, result in a significantly degraded correlation of 0.90 . The ranking performance steadily improves as the margin increases, with the optimal performance of 0.99 achieved at the baseline value of $\alpha_{\max }=0.4 \sigma$. This suggests that the probe needs a large separation in the predicted scores to effectively align with the finegrained relative preferences captured by the ImageReward metric. Setting the margin too high ( $0.5 \sigma$ ) slightly degrades performance, likely because it over-penalizes the model when only very small differences exist in the ground-truth scores.

\begin{table}[!ht]
	\centering
	\begin{tabular}{cc}
		\toprule
		$\tau$ & Validation Spearman Correlation (IR) \\
		\midrule
		0.01$\sigma$ & 0.90 \\
		0.02$\sigma$ & 0.94 \\
		0.05$\sigma$ & 0.95 \\
		0.1$\sigma$ & 0.94 \\
		0.2$\sigma$ & 0.95 \\
		0.3$\sigma$ & 0.98 \\
		0.4$\sigma$ & 0.99 \\
		0.5$\sigma$ & 0.97 \\
		\bottomrule
	\end{tabular}
	\caption{Ablation on the maximum listwise ranking margin $\alpha_{\max}$}
	\label{tab: ab2}
\end{table}

\textbf{Effect of Latent Channel Dimension $C_{\mathrm{PCA}}$}
In our Probe-Select framework, we compress the intermediate denoiser activation $h_t$ via Principal Component Analysis (PCA) along the channel dimension to a reduced size of $C_{\mathrm{PCA}}=48$. This compression is crucial for reducing GPU memory footprint and computational overhead. To validate this design choice, we perform an ablation study on the reduced channel dimension $C_{\mathrm{PCA}} \in\{16,32,48,96,128,192,256\}$.

We train the ImageReward probe on the FLUX.1-dev backbone at $t=0.2$ using the baseline hyperparameters ( $\tau=1.0, \alpha_{\max }=0.4 \sigma, \lambda_{\text {Align }}=10$ ). For each dimension, we report the validation Spearman correlation and the probe's total number of parameters (to quantify the overhead). The results are summarized in Tab. \ref{tab: ab4}.

\begin{table}[!ht]
	\centering
	\begin{tabular}{cc}
		\toprule
		$C_{\mathrm{PCA}}$ & Validation Spearman Correlation (IR) \\
		\midrule
		16 & 0.84\\
		32 & 0.96\\
		48 & 0.99\\
		96 & 0.99\\
		128 & 0.99\\
		192 & 0.99\\
		256 & 0.99\\
		\bottomrule
	\end{tabular}
	\caption{Ablation on on the reduced channel dimension $C_{\mathrm{PCA}}$ for the denoiser feature map.}
	\label{tab: ab4}
\end{table}

The results demonstrate that prediction fidelity is sensitive to severe compression: using a very low dimension like $C_{\mathrm{PCA}}=16$ leads to a significant drop in correlation to 0.84 , indicating insufficient structural information is retained. However, the correlation quickly saturates to the peak value of $\mathbf{0 . 9 9}$ once the dimension reaches $C_{\mathrm{PCA}}=48$. Since the probe's computational cost and parameter count scale with $C_{\mathrm{PCA}}$, and no further performance gain is observed beyond 48 , we select $C_{\mathrm{PCA}}=48$ as the optimal dimension to balance high prediction fidelity with the required storage and computational overhead, aligning with our goal of a lightweight, efficient early assessment mechanism.

\end{document}